\documentclass{article} % For LaTeX2e
\usepackage{iclr2027_conference,times}

\usepackage{amsmath,amsfonts,bm}

\def\eqref#1{equation~\ref{#1}}
\def\1{\bm{1}}

\DeclareMathAlphabet{\mathsfit}{\encodingdefault}{\sfdefault}{m}{sl}
\SetMathAlphabet{\mathsfit}{bold}{\encodingdefault}{\sfdefault}{bx}{n}

\usepackage{url}

\usepackage[utf8]{inputenc} % allow utf-8 input
\usepackage[T1]{fontenc}    % use 8-bit T1 fonts
\usepackage{url}            % simple URL typesetting
\usepackage{booktabs}       % professional-quality tables
\usepackage{amsfonts}       % blackboard math symbols
\usepackage{nicefrac}       % compact symbols for 1/2, etc.
\usepackage{microtype}      % microtypography
\usepackage{xcolor}         % colors
\usepackage{subfigure}
\usepackage{pgfplots}
\usepackage{pgfplotstable}
\usepackage{tikz}
\usepgfplotslibrary{groupplots}

\usepackage{graphicx}
\usepackage[font=small]{caption}

\usepackage{makecell}

\definecolor{cb_orange}{RGB}{213,94,0}
\definecolor{cb_green}{RGB}{34,136,51}
\definecolor{cbgreen}{RGB}{34,136,51}
\definecolor{sky_blue}{RGB}{204, 238, 255}
\definecolor{cb_purple}{RGB}{170, 51, 119}
\definecolor{cb_red}{RGB}{204, 51, 17}
\definecolor{cb_blue}{RGB}{0, 119, 187}
\definecolor{mydarkblue}{rgb}{0,0.08,0.45}
\definecolor{forestgreen}{RGB}{34,139,34}
\definecolor{periwinkle}{RGB}{90, 82, 235}
\definecolor{royalazure}{rgb}{0.0, 0.22, 0.66}
\definecolor{royalblue}{rgb}{0.0, 0.14, 0.4}
\usepackage[colorlinks=true,citecolor=periwinkle,linkcolor=periwinkle,urlcolor=royalazure]{hyperref}

\usepackage{tcolorbox}
\tcbuselibrary{breakable}

\usepackage{amsmath}
\usepackage{amssymb}
\usepackage{mathtools}
\usepackage{amsthm}

\usepackage[capitalize,noabbrev]{cleveref}

\usepackage{wrapfig}
\usepackage{enumitem}
\usepackage{multirow}

\theoremstyle{plain}

\theoremstyle{definition}

\theoremstyle{remark}

\definecolor{cbgreen}{RGB}{34,136,51}
\definecolor{cbblue}{RGB}{0, 119, 187}
\definecolor{cbred}{RGB}{204, 51, 17}
\definecolor{softperiwinkle}{RGB}{123, 117, 239}
\newcommand{\bluecell}[1]{\cellcolor{softperiwinkle!25}#1}
\newcommand{\planned}[1]{\textcolor{red}{#1}}
\newcommand{\placeholderfig}[2]{\IfFileExists{#1}{\includegraphics[width=#2]{#1}}{\fbox{\parbox[c][0.22\linewidth][c]{#2}{\centering\planned{Placeholder: #1}}}}}
\usepackage{bbding}          % \Checkmark, \XSolidBrush
\usepackage{colortbl}        % \cellcolor

\title{VIF-Bench: Evaluating Visual Instruction Following in Multi-Reference Image Generation}

\author{%
  \textbf{Yuta~Oshima$^{1,*}$\quad
  Masakazu~Yoshimura$^{1,*}$\quad
  Masahiro~Suzuki$^{1}$} \\
  \textbf{Yutaka~Matsuo$^{1}$\quad
  Hiroki~Furuta} \\
  \textsuperscript{1}The University of Tokyo \\
  \texttt{\{yuta.oshima, masakazu.yoshimura\}@weblab.t.u-tokyo.ac.jp}
}

\iclrfinalcopy % Uncomment for camera-ready version, but NOT for submission.
\begin{document}

\maketitle

\begin{abstract}
Recent multimodal image generation models can take multiple images and textual instructions as input, enabling reference-based generation guided not only by text but also by visual instructions such as layouts, arrows, and pose cues. 
However, existing benchmarks do not evaluate the joint setting in which multiple references must be composed under multiple and heterogeneous visual-instruction images.
To address this gap, we introduce \textbf{VIF-Bench}, a benchmark of $1{,}241$ tasks designed to assess the edge of model capabilities in this joint setting by covering: (i) multi-reference generation (up to 7) under multiple heterogeneous visual instructions (up to 6), (ii) cases where reference images can potentially compete with visual instructions (e.g., a strongly posed subject vs.\ a target pose), and (iii) controlled comparison of visual instructions with text descriptions at different levels of specificity.
Using these capabilities, we uncover three findings:
(1) models face an adherence–artifact trade-off: once models reach stronger visual instruction adherence, stronger adherence tends to coincide with more instruction artifacts in generated images,
(2) visual instruction adherence tends to be lower on tasks whose reference images carry a salient state of the controlled attribute (e.g., a neon-lit subject under a light-direction instruction), most consistently for light and wind, and
(3) for models that can understand visual instructions, it is often better to provide visual constraints directly rather than describe them in text; when using text, a moderate level of detail works better than an exhaustive description.
VIF-Bench is released as an open benchmark to establish a basis for fair comparison in controllable multi-reference image generation.
\end{abstract}

\begingroup
\renewcommand\thefootnote{*}
\footnotetext{Equal contribution}
\renewcommand\thefootnote{}
\footnotetext{Code:~\url{https://github.com/shim0114/VIF-Bench}}
\footnotetext{Benchmark:~\url{https://huggingface.co/datasets/shim0114/VIF-Bench}}
\endgroup

\section{Introduction}
\label{sec:intro}

Recent image generation models, built on large multimodal LLM backbones, have advanced to a stage where they can generate not only from textual prompts but also by understanding multiple reference images and visual control marks~\citep{google2025nanobanana, google2025nanobananapro, openai2025gpt4oimage, openai2025gptimage1_5, wu2025qwenimagetechnicalreport}. 
Both multi-reference image generation~\citep{wu2025omnigen2, xia2025dreamomni2, oshima2026multibanana, zhang2026rcedit, huang2026scaling}, which recomposes subjects and backgrounds in new contexts, and visual-instruction-following generation~\citep{chen2025multiref, zhang2026vibe-benchmark, xia2025dreamomni3, ghazanfari2025spotedit}, which interprets control marks such as layouts, masks, and arrows, share a common foundation: they both rely on the visual understanding capability of VLMs to interpret visual inputs at a semantic rather than pixel level.
In practical workflows, users want to specify not only \emph{what} to generate but also \emph{how} to compose it. 
Given multiple references, users may specify placement via layouts, 3D orientation cues (rendered as pyramids), and global effects such as wind or lighting via arrows. Such control is directly relevant to a wide range of applications, including advertising~\citep{inoue2023layout, morita2025tkg}, virtual try-on~\citep{choi2024improving, cvpr2026garments2look}, and content creation~\citep{ruiz2022dreambooth, zhou2024migc, xu2026contextgen}.

\begin{figure}[t]
\centering
\includegraphics[width=\linewidth]{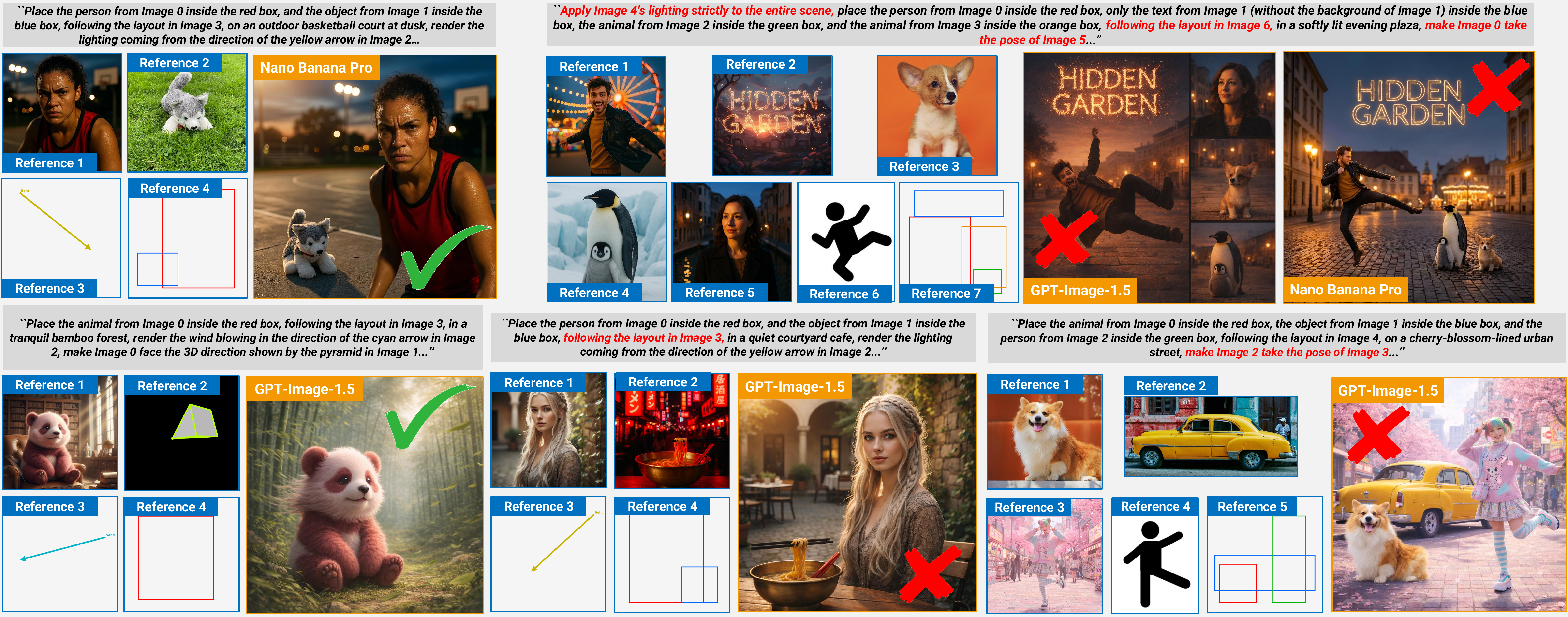}
% \vspace{-0.15in}
\caption{
Overview of \textbf{VIF-Bench}.
VIF-Bench covers challenges in multi-reference, multi-visual-instruction settings, including multiple references (up to 7) with multiple visual instructions (up to 6) and potential reference--visual-instruction conflicts, such as lighting conflicts (bottom middle) or pose conflicts (bottom right).
VIF-Bench covers various types of visual instructions, including layout, 3D orientation, pose, and wind/light direction.
}
% \vspace{-0.1in}
\label{fig:teaser}
\end{figure}

However, existing benchmarks evaluate multi-reference generation and visual control largely in isolation, and therefore do not fully capture the challenges that arise when the two are combined. 
They do not evaluate how multiple independent subject references interact with multiple visual instruction images, especially when reference images interfere with the instructions. 
Moreover, a further challenge arises from how these visual controls are represented in recent multimodal generators. 
Unlike earlier approaches that rely on dedicated conditioning modules~\citep{zhang2023controlnet}, recent models~\citep{google2025nanobanana, openai2025gpt4oimage, wu2025qwenimagetechnicalreport} interpret layouts, arrows, and orientation cues directly as image inputs. 
This unified image-based interface introduces failure modes specific to visual instructions: models may reproduce the instruction marks themselves in the output or fail to preserve the intended constraint. 
Existing benchmarks do not capture these failure modes in settings with multiple references and heterogeneous visual instructions.

To evaluate this setting, we introduce \textbf{VIF-Bench}, which jointly assesses multi-reference composition and heterogeneous visual-instruction following (\autoref{fig:teaser}). 
VIF-Bench comprises $1{,}241$ tasks designed to assess the edge of model capabilities in this joint setting by covering: (i) multi-reference generation (up to 7) under multiple heterogeneous visual instructions (up to 6), (ii) cases where reference images can potentially compete with visual instructions (e.g., a strongly posed subject vs.\ a target pose), and (iii) controlled comparison of visual instructions with text descriptions at different levels of specificity.
Using VIF-Bench, we identify three major findings: 
(1) models face an adherence–artifact trade-off: once models reach stronger visual instruction adherence, stronger adherence tends to coincide with more instruction artifacts,
(2) visual instruction adherence tends to be lower on tasks whose reference images carry a salient state of the controlled attribute (e.g., a neon-lit subject under a light-direction instruction), most consistently for light and wind, and
(3) for models that can understand visual instructions, it is often better to provide visual constraints directly rather than describe them in text; when using text, a moderate level of detail works better than an exhaustive description.
These findings reveal limitations specific to jointly satisfying multiple references and heterogeneous visual constraints. 
We release VIF-Bench as an open benchmark for evaluating controllable multi-reference image generation.

\section{Related Works}
\label{sec:related}

\subsection{Controllable Text-to-Image Generation}
\label{sec:rel_controllable}

Diffusion models have achieved state-of-the-art performance in high-fidelity image synthesis~\citep{sohl-dickstein15, ho2020ddpm}. 
Large pretrained diffusion models with textual conditioning, such as Stable Diffusion~\citep{rombach2022ldm, podell2024sdxl, esser2024sd3} and FLUX~\citep{flux2024, labs2025flux1kontext}, form the foundation of modern text-to-image generation. 
To improve controllability, prior work has introduced additional conditioning channels, including spatial maps such as edges, depth, segmentation, and boxes~\citep{zhou2024migc, zhang2023controlnet, li2023gligen, xu2026contextgen}, as well as identity-preserving reference conditioning through fine-tuning or adapters~\citep{ruiz2022dreambooth, ye2023ip-adapter, mou2023t2i-adapter}.
Recent multimodal image generation models further extend this direction by jointly processing text and image inputs within a unified framework.
For example, closed-source systems such as GPT-Image~\citep{openai2025gpt4oimage, openai2025gptimage1_5, openai2026gptimage2} and Nano Banana~\citep{google2025nanobanana, google2025nanobananapro, google2025nanobanana2} enable integrated image generation and editing from mixed text-image prompts.
Similarly, open-source models such as Qwen-Image~\citep{wu2025qwenimagetechnicalreport} and FLUX Kontext~\citep{labs2025flux1kontext} demonstrate high-quality and flexible image generation and editing, and numerous unified generative models continue to emerge~\citep{wu2025qwenimagetechnicalreport, wu2025omnigen2, xia2025dreamomni2, wu2025less, deng2025bagel, xie2024showo}.
These advances show that image generation models are becoming increasingly capable of interpreting heterogeneous inputs, including reference images, styles, spatial layouts, and other visual cues~\citep{chen2025multiref, xia2025dreamomni3, zhang2026vibe-benchmark}.
VIF-Bench asks whether such models can actually follow these visual instructions when multiple reference subjects must be composed together.

\begin{table*}[t]
\centering
\small
\caption{
Comparison among major benchmarks for reference-based image generation,
editing, and visual-instruction following.
\textit{\#Refs} denotes the maximum number of reference images provided
within a task, and \textit{\#VIs} denotes the maximum number of explicit visual-instruction images that can be combined within a single task.
\textit{Reference--VI Conflict} indicates whether the benchmark explicitly
identifies potential conflict cases in which attributes implied by a reference image may compete with the corresponding visual instruction.
VIF-Bench jointly evaluates multiple references and heterogeneous visual
instructions while explicitly identifying reference--visual-instruction
conflict.
\textsuperscript{$\dagger$}DreamOmni3 only studies scribble-guided editing.
}
\label{tab:benchmarks}
% \vspace{-0.1in}
\scalebox{0.8}{
\begin{tabular}{lccccl}
\toprule
\textbf{Benchmark}
& \textbf{\#Size}
& \textbf{\#Refs}
& \textbf{\#VIs}
& \textbf{Reference--VI Conflict}
& \textbf{Metrics} \\
\midrule

\multicolumn{6}{l}{\textit{Without visual instructions}} \\

DreamBooth~\citep{ruiz2022dreambooth}
& 75
& 1
& --
& \textcolor{cb_red}{\XSolidBrush}
& CLIP, DINO \\

OmniContext~\citep{wu2025omnigen2}
& 400
& 3
& --
& \textcolor{cb_red}{\XSolidBrush}
& GPT (3 dim.) \\

DreamOmni2~\citep{xia2025dreamomni2}
& 319
& 4
& --
& \textcolor{cb_red}{\XSolidBrush}
& Gemini, Doubao~\citep{bytedance2025doubao} \\

MultiBanana~\citep{oshima2026multibanana}
& 3{,}769
& 8
& --
& \textcolor{cb_red}{\XSolidBrush}
& GPT, Gemini (5 dim.) \\

\midrule
\multicolumn{6}{l}{\textbf{\textit{With visual instructions}}} \\

MultiRef~\citep{chen2025multiref}
& 1{,}990
& 6
& 1
& \textcolor{cb_red}{\XSolidBrush}
& GPT (3 dim), modality-specific metrics \\

VIBE~\citep{zhang2026vibe-benchmark}
& 1{,}034
& 1
& 1
& \textcolor{cb_red}{\XSolidBrush}
& GPT (3 dim.) \\

DreamOmni3~\citep{xia2025dreamomni3}
& 731
& 4
& $2^\dagger$
& \textcolor{cb_red}{\XSolidBrush}
& Gemini, Doubao~\citep{bytedance2025doubao} \\

\bluecell{\textbf{VIF-Bench (Ours)}}
& \bluecell{1{,}241}
& \bluecell{7}
& \bluecell{6}
& \bluecell{\textbf{\textcolor{cb_green}{\Checkmark}}}
& \bluecell{GPT, Gemini, Qwen (6 dim.)} \\

\bottomrule
\end{tabular}
}
% \vspace{-0.1in}
\end{table*}

\subsection{Benchmarks for Reference-Based Generation}
\label{sec:rel_benchmarks}

Benchmarks for reference-based image generation trace back to DreamBooth~\citep{ruiz2022dreambooth}, which evaluates subject-driven generation conditioned on a single subject.
Subsequent studies have extended this setting to multiple references, introducing benchmarks for multi-reference generation~\citep {zong2024easyref, sushko2025realedit, wu2025omnigen2, xia2025dreamomni2, oshima2026multibanana, chen2026mibe}.
Other studies have extended this setting to visual instruction following.
MultiRef~\citep{chen2025multiref} supports multiple reference images, but each task uses at most a single visual-instruction image. VIBE~\citep{zhang2026vibe-benchmark} covers regions, morphological cues such as pose and orientation, and arrows, and can combine multiple visual operations within a single annotated image. DreamOmni3~\citep{xia2025dreamomni3} supports up to two visual-instruction images, but focuses only on scribble-based control. 
Existing benchmarks therefore do not evaluate the joint setting targeted in this work, where multiple references must be composed simultaneously under several heterogeneous visual-instruction images.
See Appendix~\ref{sec:prior_work_comparison_appendix} for further discussion.

\section{VIF-Bench}
\label{sec:ifbanana}

\begin{figure}[t]
    \centering
    \begin{minipage}{0.54\linewidth}
        \centering
        \includegraphics[width=\linewidth]{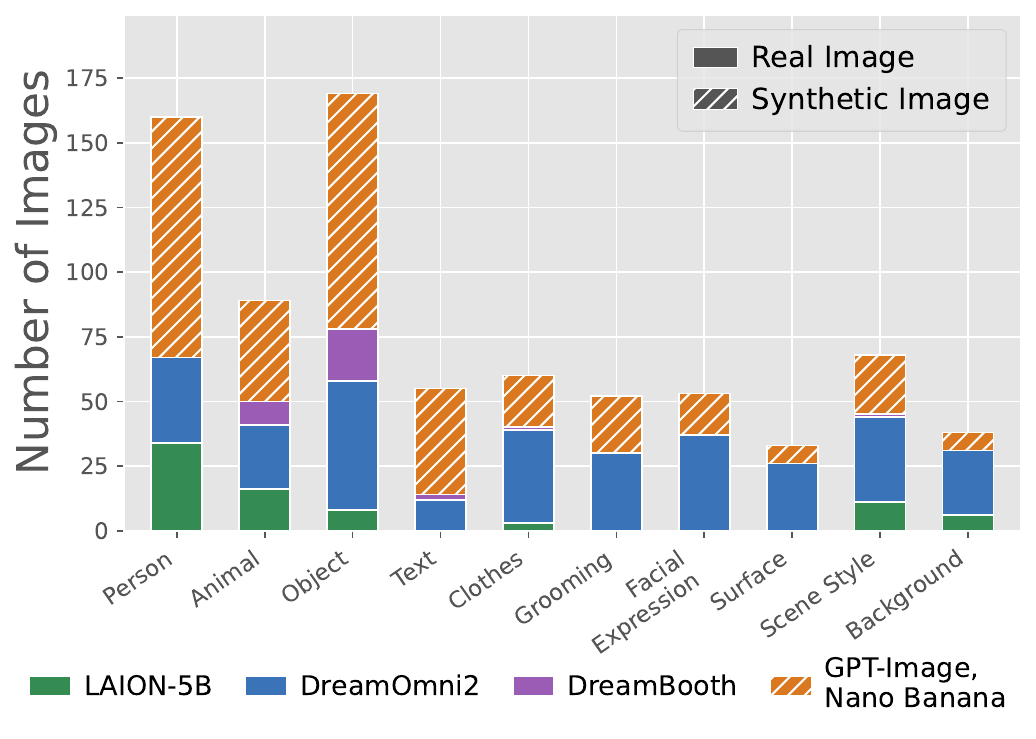}
    \end{minipage}\hfill
    \begin{minipage}{0.45\linewidth}
        \centering
        \includegraphics[width=\linewidth]{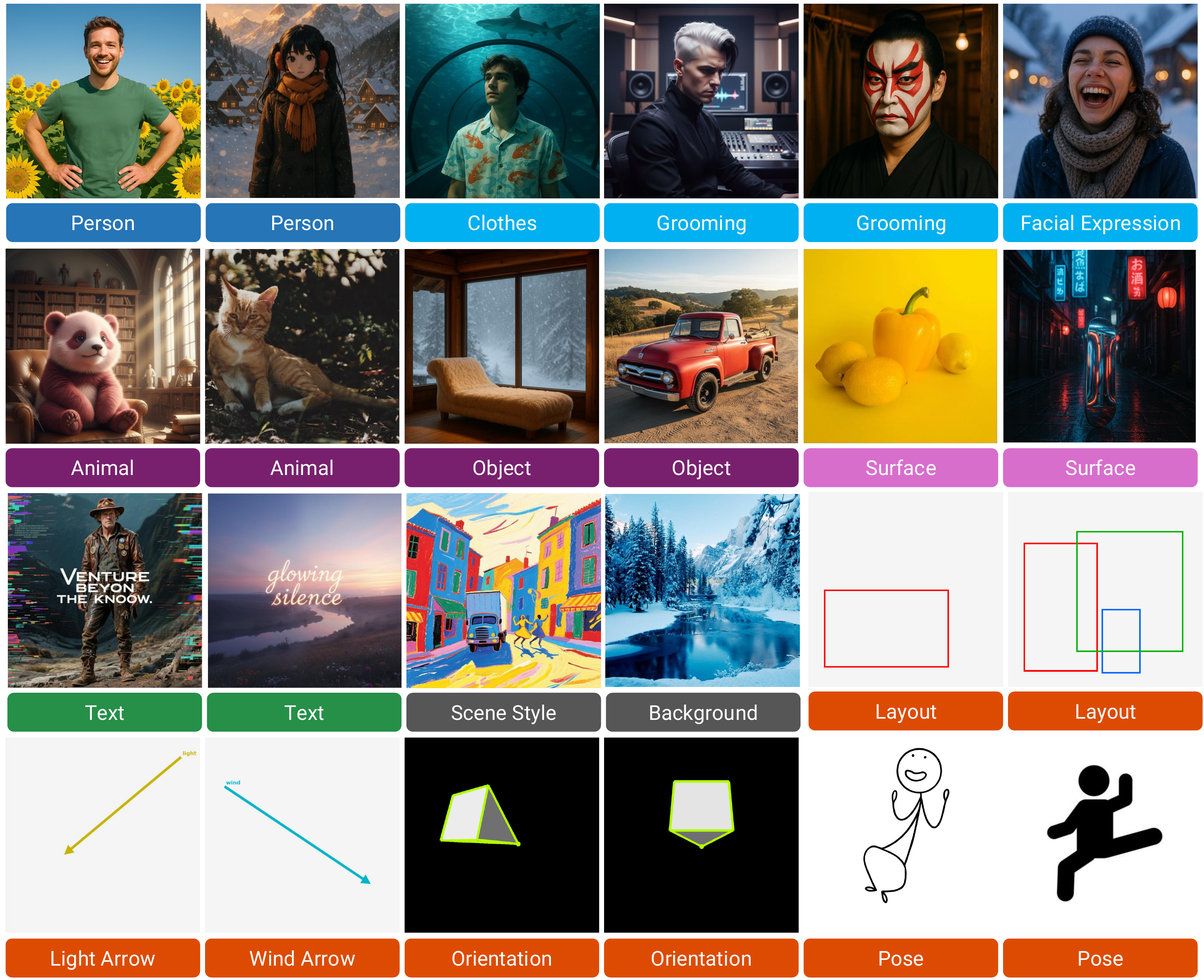}
    \end{minipage}
    % \vspace{-0.1in}
    \caption{(\textbf{Left}) Per-category counts of reference images in VIF-Bench, broken down into three families: Main Reference (i.e., person, animal, object, text), Sub Reference (i.e., clothes, grooming, facial expression, surface), and Scene Context (scene style, background). 
    Combining multiple real-image datasets with synthetic images from image-generation models lets us populate not only the common Main Reference categories but also Sub Reference categories that are scarce in general-purpose datasets (e.g., grooming, facial expression, and object surface), yielding a balanced pool across the taxonomy.
    (\textbf{Right}) Examples of reference images and visual instructions sampled from VIF-Bench. We prepare various types of visual instructions, such as layout, light arrow, wind arrow, orientation, and human pose.}
    \label{fig:image_source_stats}
    % \vspace{-0.15in}
\end{figure}

\subsection{Construction of VIF-Bench}
\label{sec:construction}

\noindent\textbf{Image Collection.}~~
We construct the reference-image pool from both real and synthetic images.
Real images are collected from LAION-5B~\citep{schuhmann2022laion}, DreamBooth~\citep{ruiz2022dreambooth}, and DreamOmni2~\citep{xia2025dreamomni2}, while synthetic images are generated by Nano Banana~\citep{google2025nanobanana} and GPT-Image-1~\citep{openai2025gpt4oimage}.
Combining multiple data sources covers both common subject references and scarce attribute-specific references, such as hairstyle, makeup, facial expression, and object surface.

\noindent\textbf{Category Classification.}~~
Following prior multi-reference image-generation benchmarks~\citep{xia2025dreamomni2,oshima2026multibanana}, we hierarchically classify the collected images into three groups: Main Reference, Sub Reference, and Scene Context.
GPT-5~\citep {openai2025gpt5} first automatically labels the images, which we then manually verify and correct.
Main Reference denotes the primary subject and consists of \textit{person}, \textit{animal}, \textit{object}, and \textit{text}.
Objects are further divided into \textit{versatile}, \textit{indoor-only}, and \textit{outdoor-only} according to their plausible placement environments.
Sub Reference specifies attributes of a main subject: for people, we consider \textit{clothes}, \textit{grooming} (e.g., hairstyle and makeup), and \textit{facial expression}; for objects, we consider \textit{surface} properties such as color and material.
Scene Context consists of \textit{scene style} and \textit{background}, which specify the overall appearance of the generated scene.
This hierarchy prevents semantically inconsistent task construction, such as assigning a facial-expression reference to an object.

\noindent\textbf{Task Construction.}~~
Each task is constructed by combining hierarchically organized reference
images with visual instructions.
The input conditions are broadly divided into per-subject conditions,
associated with individual main subjects, and global conditions, associated
with the scene as a whole.
The former include sub-references, pose references, and 3D orientation
pyramids, while the latter include style modifiers, backgrounds, and
directional arrows.
Concretely, we (1) sample
$N_{main}\!\in\!\{1,2,3,4\}$ main subjects together with their associated
references, (2) ask GPT-5~\citep{openai2025gpt5} to propose a bounding-box
layout, per-subject 3D orientations, and wind/light arrow directions, which
are then rendered as visual-instruction images, and (3) generate a textual
instruction from the resulting reference structure using a deterministic
template.
These visual instructions are constructed to precisely specify
the target spatial and geometric conditions and to enable unambiguous
evaluation of instruction following, allowing task construction to scale to
a large benchmark.
To ensure task validity, all constructed tasks first undergo automatic
consistency checks and are then verified by Gemini~\citep{geminiteam2023gemini}
and human annotators; unnatural, ambiguous, or inconsistent tasks are
corrected or removed.
Pose references, orientation cues, and directional arrows follow
visual-control formulations used in prior visual-instruction benchmarks
~\citep{zhang2026vibe-benchmark}.
Bounding-box layouts are likewise a standard spatial-control interface
widely used in prior work~\citep{li2023gligen,zhou2024migc}.
Thus, while VIF-Bench procedurally generates visual instructions in a
controlled manner for reliable evaluation, the instruction representations
themselves follow established and practical visual-control paradigms.
We additionally construct three text-converted variants for a stratified 200-task subset,
enabling controlled comparison of instruction modality and specificity.
Further construction details are provided in
Appendix~\ref{sec:appendix-vi-text}.

\noindent\textbf{Conflict Detection.}~~
We define \emph{conflict tasks} as cases in which a reference image contains a salient state along an attribute that is also controlled by the corresponding visual instruction, creating the potential for the reference-implied state to compete with the requested control. 
For example, if a person in the reference image is strongly side-lit while a light arrow specifies frontal illumination, the model must override the lighting implied by the reference and follow the visual instruction.
We identify conflict along four axes:

\begin{itemize}[leftmargin=0.5cm,topsep=0pt,itemsep=0pt]
  \item \textbf{Orientation conflict.}~~
  A potential conflict is declared when the subject in the reference image has a clear facing direction, and the target orientation specified by the visual instruction departs substantially from it. Concretely, we target cases where the yaw difference between the reference and the instruction is large, or where the left/right facing direction is flipped.
  \item \textbf{Light conflict.}~~
  A potential conflict is declared when the reference image contains a distinctive light source that strongly shapes the scene's appearance—such as neon or moonlight—rather than ordinary daylight or uniform illumination, and the task includes a light visual instruction.
  \item \textbf{Wind conflict.}~~
  We declare a potential conflict when the reference image contains elements whose appearance changes substantially under wind, such as hair or fabric, and the task includes a wind visual instruction.
  \item \textbf{Pose conflict.}~~
  We declare a potential conflict when a person in the reference image is in a clear, distinctive pose and a pose visual instruction is assigned to that person.
\end{itemize}

We do not define a conflict criterion for layout because layout specifies the spatial arrangement of the overall scene rather than an intrinsic attribute of a reference image.
We first automatically label the reference-specific attributes required for these judgments with VLMs, then manually verify and correct them.
By design, these labels identify \emph{potential} conflicts based on reference content: orientation can be compared directly against the target via a yaw estimate, whereas for light, wind, and pose we flag cases where the controlled attribute is salient in the reference and must be re-rendered to satisfy the visual instruction.
Further details for benchmark construction are shown in Appendix~\ref{sec:appendix-task-construction}.

\subsection{Statistics in VIF-Bench}
\label{sec:statistics}

\noindent\textbf{Image Statistics.}~~
The reference-image pool consists of 777 images (\autoref{fig:image_source_stats}; \textbf{Left}).
Main Reference contains 160 person images, 89 animal images, 169 object images, and 55 text images.
Sub Reference contains 60 clothes images, 52 grooming images, 53 facial-expression images, and 33 surface images.
Scene Context contains 68 scene-style images and 38 background images.
For text typography, such as language and font, we reuse the main text-reference pool.
By source, 78 images are from LAION-5B~\citep{schuhmann2022laion}, 307 from DreamOmni2~\citep{xia2025dreamomni2}, 33 from DreamBooth~\citep{ruiz2022dreambooth}, and 359 are synthetic images generated by Nano Banana~\citep{google2025nanobanana} and GPT-Image-1~\citep{openai2025gpt4oimage}, resulting in a mixture of real and synthetic images.

\begin{figure}[t]
    \centering
    \begin{minipage}[t]{0.21\linewidth}
        \centering
        \includegraphics[width=\linewidth]{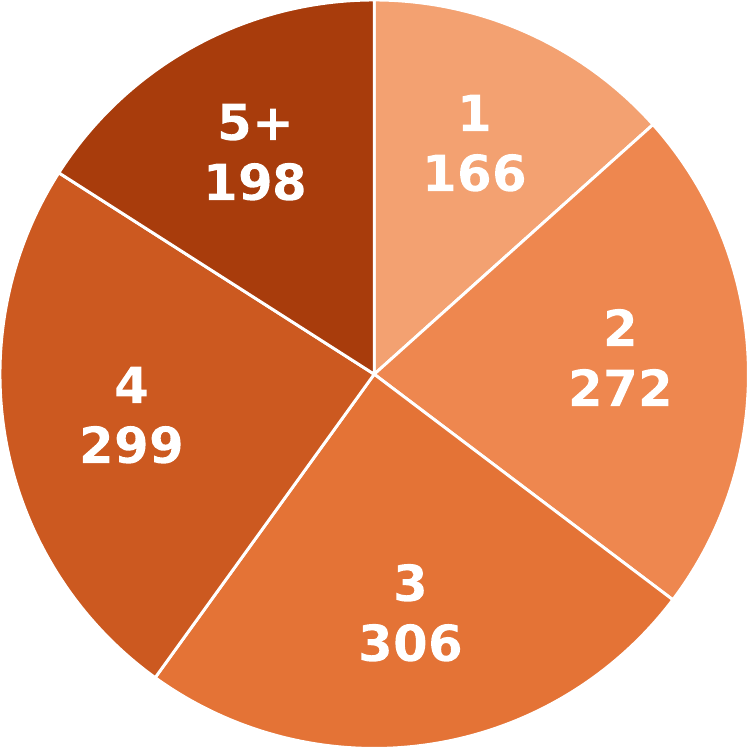}
        \small (a) References per task
    \end{minipage}
    \begin{minipage}[t]{0.21\linewidth}
        \centering
        \includegraphics[width=\linewidth]{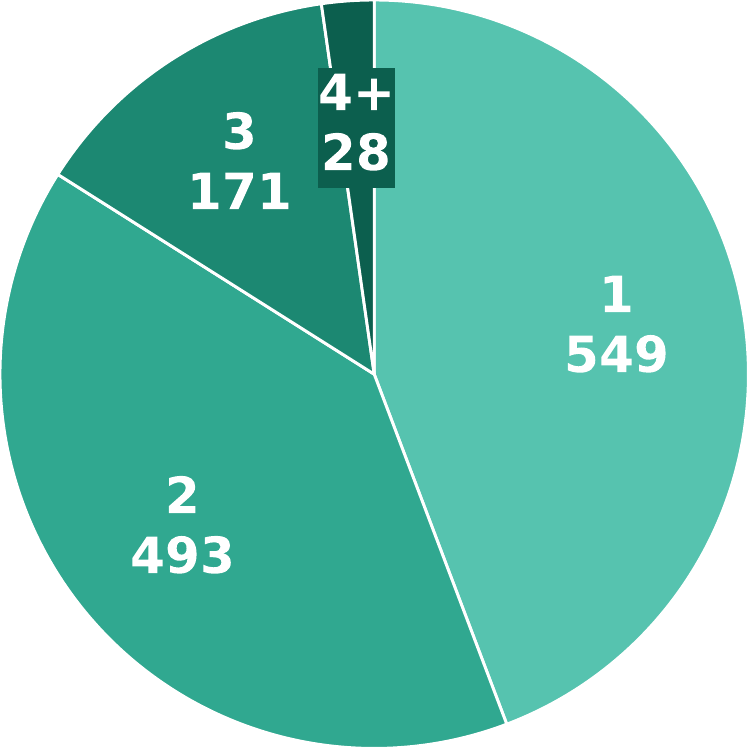}
        \small (b) Visual instructions
    \end{minipage}
    \begin{minipage}[t]{0.265\linewidth}
        \centering
        \includegraphics[width=\linewidth]{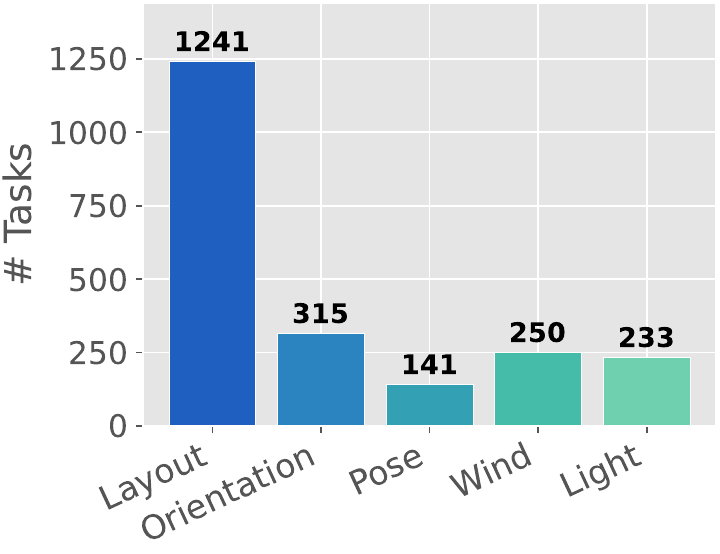}
        \small (c) Adoption counts by kind
    \end{minipage}
    \begin{minipage}[t]{0.26\linewidth}
        \centering
        \includegraphics[width=\linewidth]{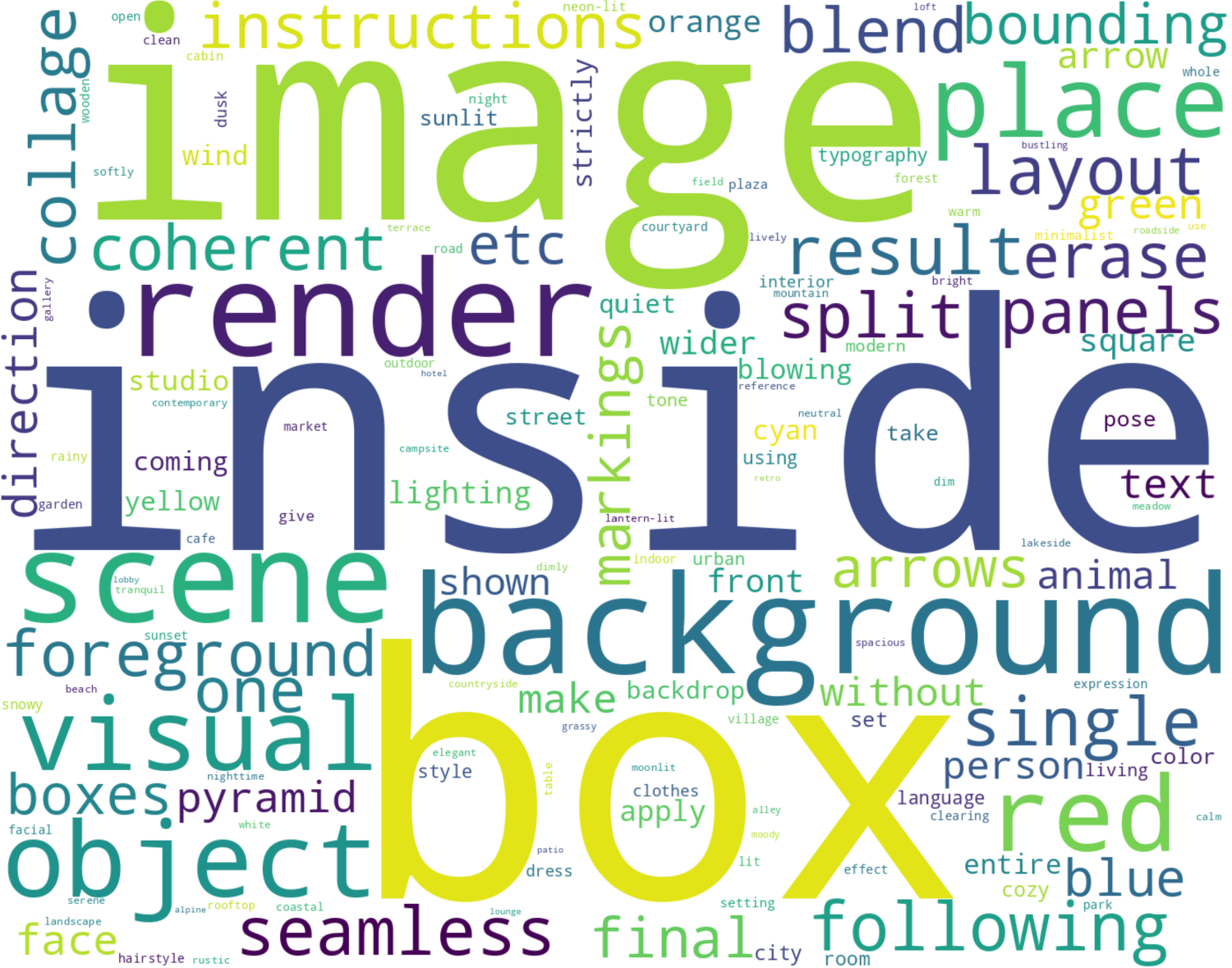}
        \small (d) Word cloud
    \end{minipage}

    \caption{Statistics of the VIF-Bench evaluated set (1{,}241 tasks).
    \textbf{(a)} Distribution of the number of reference images per task.
    \textbf{(b)} Distribution of the number of visual-instruction images per task.
    \textbf{(c)} Adoption counts of visual instructions by kind.
    \textbf{(d)} Word cloud of the textual instructions. It primarily consists of terms that describe a wide range of object categories and words indicating spatial directions.
    }
    \label{fig:task_stats}
    % \vspace{-0.1in}
\end{figure}

\noindent\textbf{Task Statistics.}~~
The evaluated set consists of 1{,}241 tasks. As shown in \autoref{fig:task_stats} (\textbf{a, b}), the number of reference images and visual instructions per task is broadly distributed: tasks contain between 1 and 7 reference images (166, 272, 306, 299, 161, 32, and 5 tasks, respectively) and between 1 and 6 visual-instruction images (549, 493, 171, 23, 4, and 1 tasks, respectively). 
Broken down by type (\autoref{fig:task_stats}; \textbf{c}), layout is adopted in all 1{,}241 tasks and serves as the common backbone, with orientation (315), wind (250), light (233), and pose (141) co-occurring on top of it.
We include layout in every task by design because spatial placement is a fundamental component of multi-reference composition and bounding boxes provide a simple, widely used control interface~\citep{inoue2023layout, feng2024layoutgpt}, while the remaining visual instructions are layered on top as optional controls.
% Further statistics are shown in Appendix~\ref{sec:appendix-further-stats}.

\paragraph{Conflict Statistics.}
Among the 1,241 tasks, 438 contain at least one conflict between a reference-image attribute and its visual instruction, whereas 803 contain no such conflict.
At the instruction-type level, orientation conflict cases occur in 106 of the 315 tasks containing an orientation instruction (33.7\%), pose conflict in 82 of 141 pose tasks (58.2\%), wind conflict in 153 of 250 wind tasks (61.2\%), and light conflict in 172 of 233 light tasks (73.8\%).
Because a single task may contain multiple conflict types, these categories are not exclusive.

\subsection{Evaluation Setting}
\label{sec:evaluation}

Because large-scale human evaluation is prohibitively expensive, we adopt VLM-based evaluation, which is widely used in recent multimodal-generation benchmarks~\citep{ku2023viescore, na2024boost, oshima2025inference}.
Gemini 2.5 Flash~\citep{geminiteam2023gemini} and GPT-5~\citep{openai2025gpt5} independently evaluate all generated images, and we report their average score.
Following prior reference-based image-generation benchmarks~\citep{ye2025imgedit, wu2025omnigen2, oshima2026multibanana}, we retain the core evaluation dimensions of instruction following, reference consistency, and overall image quality.
We further separate overall image quality into Scene Coherence and Visual Quality for more fine-grained assessment.
Accordingly, each image is evaluated on a 10-point scale along six criteria: Text Instruction Following, Reference Consistency, Vision Instruction Adherence, Visual Instruction Cleanliness, Scene Coherence, and Visual Quality.
Text Instruction Following and Reference Consistency assess adherence to the textual instruction and preservation of reference subjects, while Scene Coherence and Visual Quality assess overall scene consistency and perceptual quality.
To capture failure modes specific to image-based visual control in recent multimodal image generation models~\citep{google2025nanobanana, openai2025gpt4oimage, wu2025qwenimagetechnicalreport}, we separately evaluate Vision Instruction Adherence and Visual Instruction
Cleanliness.
The former measures whether the generated image satisfies the constraints specified by the visual instructions, while the latter measures whether instruction elements such as layout boxes and arrows remain in the output.

\section{Experiments}
\label{sec:experiments}

We evaluate a representative set of state-of-the-art multi-reference image
generators on the full $1{,}241$-task VIF-Bench benchmark.
The closed-source models include
Nano Banana Pro~\citep{google2025nanobananapro},
Nano Banana~\citep{google2025nanobanana},
GPT-Image-1.5~\citep{openai2025gptimage1_5}, and
GPT-Image-1~\citep{openai2025gpt4oimage},
while the open-source models include
DreamOmni2~\citep{xia2025dreamomni2},
FLUX.1 Kontext~\citep{labs2025flux1kontext},
Qwen-Image-Edit-2511, and
Qwen-Image-Edit-2509~\citep{wu2025qwenimagetechnicalreport}.
Each model receives the reference sequence and structured prompt defined in
Section~\ref{sec:construction}.
Each generated image is independently evaluated by
Gemini 2.5 Flash~\citep{geminiteam2023gemini} and
GPT-5~\citep{openai2025gpt5} using the six criteria described in
Section~\ref{sec:evaluation}, and we report the average of the two judges.
See Appendix~\ref{sec:qwen_results} for the results of Qwen3-VL~\citep{bai2025qwen3vl} as judge, 
Appendix~\ref{sec:additional_qualitative} for additional qualitative results, and Appendix~\ref{sec:appendix_further_resutls} for the detailed discussion about evaluation.

\subsection{Overall and Per-Visual-Instruction Evaluation}
\label{sec:exp_per_vi}

\begin{figure}[t]
    \centering
    % \vspace{-0.1in}
    \begin{minipage}{0.265\linewidth}
        \centering
        \includegraphics[width=\linewidth]{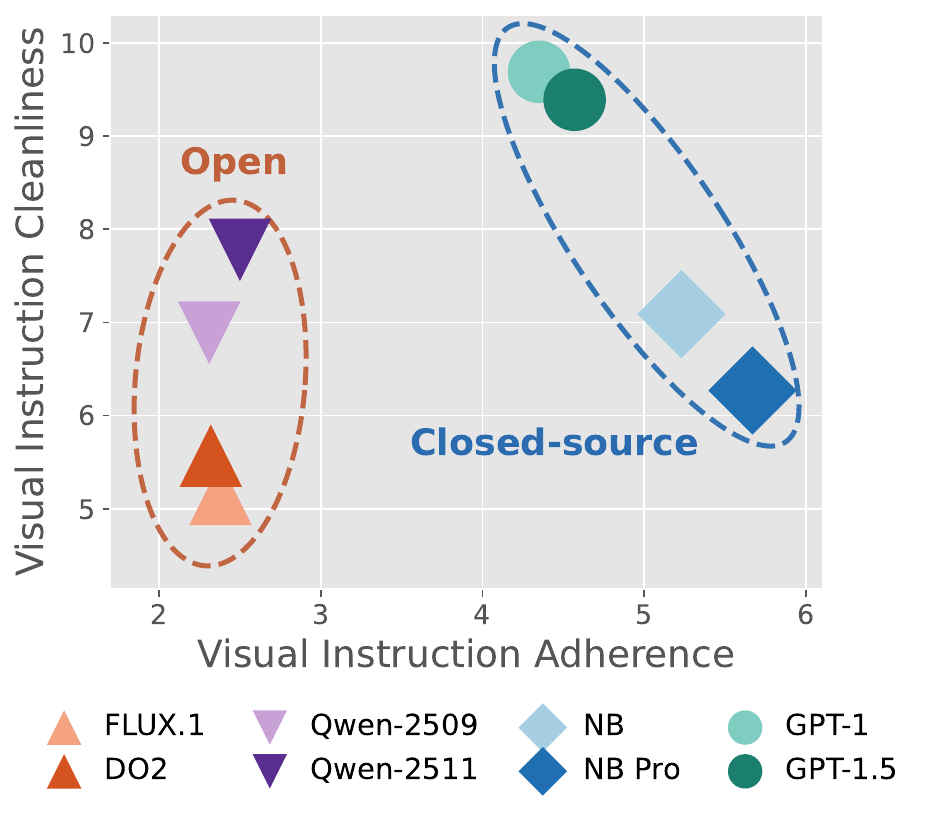}
    \end{minipage}\hfill
    \begin{minipage}{0.66\linewidth}
        \centering
        \includegraphics[width=\linewidth]{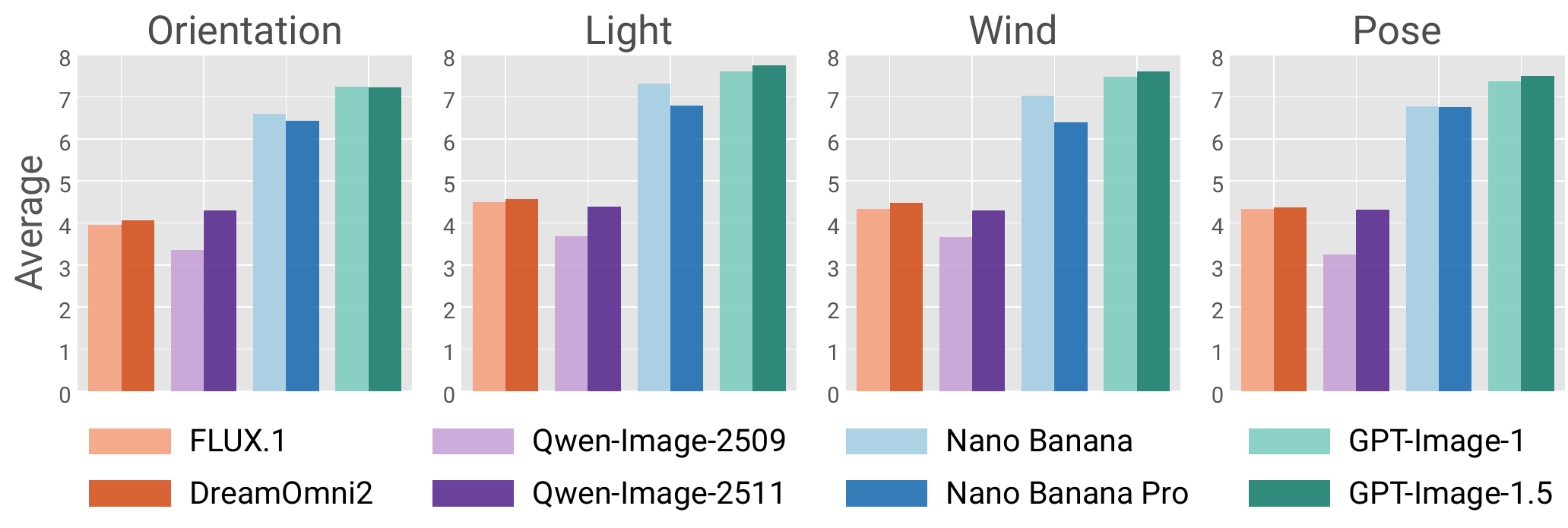}
    \end{minipage}
    % \vspace{-0.1in}
    \caption{(\textbf{Left}) Two-regime adherence–artifact pattern across generators.  The models separate into two distinct clusters (dashed ellipses): open-weight models group tightly at low adherence, whereas closed models form a separate cluster at higher adherence. 
    Within the closed cluster, however, stronger adherence tends to coincide with reduced cleanliness, suggesting a trade-off between the two objectives.
    (\textbf{Right}) Average VIF-Bench score on tasks containing each visual-instruction type.
    Visual instructions adopted in VIF-Bench are sensitive to improvements within model families.}
    \label{fig:scatter_and_per_visual_instruction}
    % \vspace{-0.15in}
\end{figure}

\begin{figure*}[t]
    \centering
    % \placeholderfig{figures/failure_mode.pdf}{\linewidth}
    \includegraphics[width=0.98\linewidth]{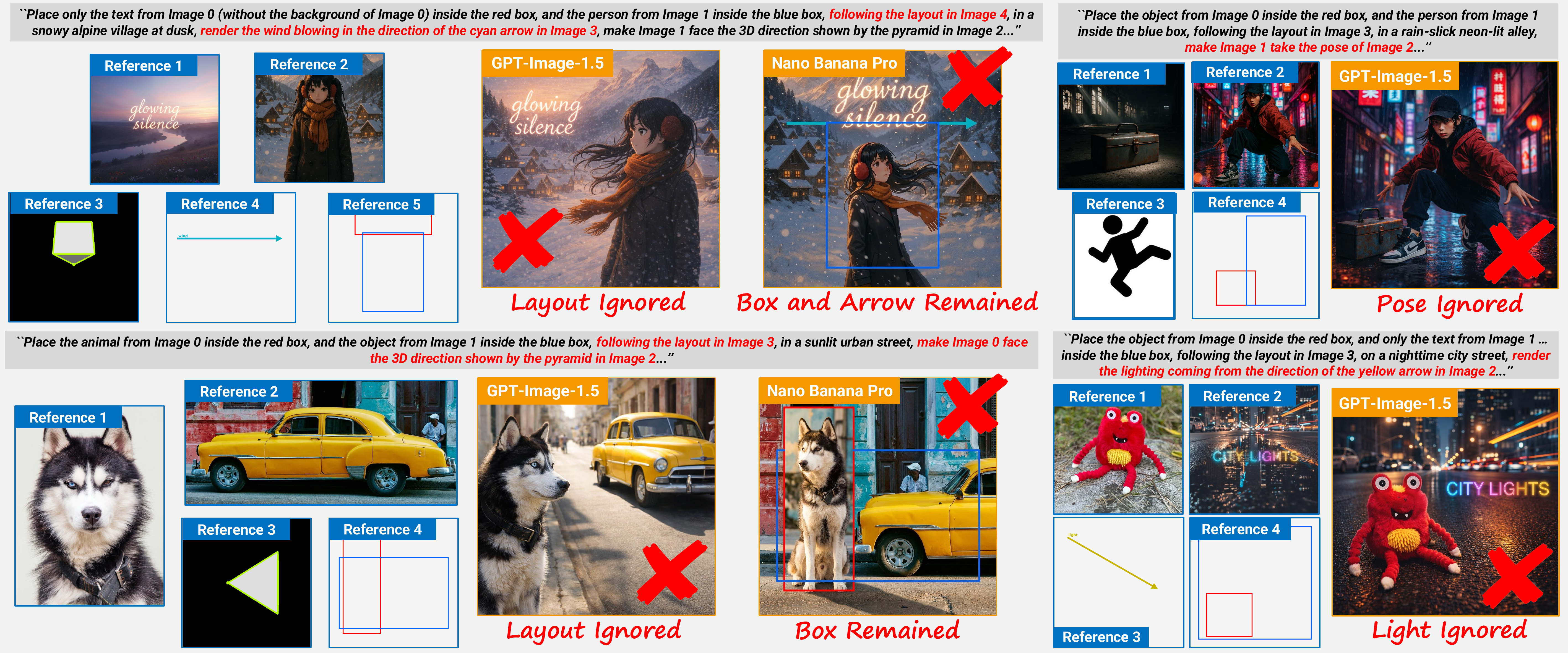}
    % \vspace{-0.2in}
    \caption{
    % Representative failure modes. 
    Qualitative examples of VIF-Bench.
    (\textbf{Left}) The adherence–artifact trade-off among closed models: layout drift in GPT-Image-1.5 and residual instructions in Nano Banana Pro. 
    (\textbf{Right}) Reference--visual-instruction conflict in pose or lighting results in ignorance of visual instructions.
    }
    \label{fig:qualitative}
    % \vspace{-0.05in}
\end{figure*}

\autoref{tab:per_vi} reports the overall performance of each generator
across the six evaluation criteria.
% GPT-Image-1.5 achieves the highest average score of $7.60$, followed by
% GPT-Image-1 at $7.55$.
Closed-source models outperform open-source models overall.
However, Visual Instruction Adherence remains challenging even for the
strongest models, with the best score reaching only $5.67$.
This indicates that satisfying visual instructions, beyond
preserving multiple reference subjects, remains a major challenge for
current generators.
Separating Visual Instruction Adherence from Visual Instruction Cleanliness reveals a two-regime adherence–artifact pattern (\autoref{fig:scatter_and_per_visual_instruction}; \textbf{Left}). 
Open-weight models generally score low on Adherence, indicating that they often fail to follow the visual instructions in the first place. Closed models form a higher-adherence regime, but within this group, models with stronger Adherence tend to show lower Cleanliness. 
The GPT-Image models achieve high Cleanliness but relatively lower Adherence, whereas the Nano Banana models show stronger Adherence with more residual instruction marks. 
This closed-model adherence–artifact trade-off is also qualitatively illustrated in \autoref{fig:qualitative} (\textbf{Left}).

\autoref{fig:scatter_and_per_visual_instruction} (\textbf{Right}) reports the average score on tasks containing each visual-instruction type. Across all visual-instruction types, closed-source models consistently outperform open-source models, indicating a capability gap across different forms of visual control. At the same time, visual instructions in VIF-Bench are sensitive to improvements within model families: in most cases, they capture gains from model fine-tuning or newer versions, such as from FLUX.1 Kontext to DreamOmni2, from Qwen-Image-Edit-2509 to 2511, and from GPT-Image-1 to 1.5. Together, these trends demonstrate that VIF-Bench is sufficiently discriminative to capture both broad capability gaps across model families and incremental improvements across successive model variants.

\begin{table}[t]
\centering
\small
\caption{Overall VIF-Bench scores for each generator under the six evaluation criteria. We additionally report results for the agentic multi-step variants of GPT-Image-1.5 and Nano Banana Pro.}
% \vspace{-0.1in}
\label{tab:per_vi}
\scalebox{0.79}{
\begin{tabular}{l c c c c c c c}
\toprule
\textbf{Model} &
\makecell{\textbf{Text Instruction}\\\textbf{Following}} &
\makecell{\textbf{Reference}\\\textbf{Consistency}} &
\makecell{\textbf{Visual Instruction}\\\textbf{Adherence}} &
\makecell{\textbf{Visual Instruction}\\\textbf{Cleanliness}} &
\makecell{\textbf{Scene}\\\textbf{Coherence}} &
\makecell{\textbf{Visual}\\\textbf{Quality}} &
\textbf{Avg.} \\
\midrule
GPT-Image-1.5 & 6.79 & 8.12 & 4.57 & 9.39 & 7.84 & 8.88 & \textbf{7.60} \\
 ~ + multi-step & 6.53 & 7.29 & 4.43 & \textbf{9.80} & \textbf{8.24} & 8.89 & 7.53 \\
\midrule
Nano Banana Pro & 6.07 & 8.27 & \textbf{5.67} & 6.27 & 7.01 & 8.48 & 6.96 \\
 ~ + multi-step & \textbf{6.85} & 7.37 & 5.09 & 8.37 & 8.03 & 8.69 & 7.40 \\
\midrule
GPT-Image-1 & 6.51 & 7.84 & 4.35 & 9.69 & 7.99 & \textbf{8.90} & 7.55 \\
Nano Banana & 6.40 & \textbf{8.39} & 5.23 & 7.09 & 7.13 & 8.55 & 7.13 \\
Qwen-Image-2511 & 3.01 & 3.38 & 2.50 & 7.78 & 5.62 & 7.33 & 4.94 \\
Qwen-Image-2509 & 2.54 & 2.85 & 2.31 & 6.89 & 4.31 & 4.68 & 3.93 \\
DreamOmni2 & 2.86 & 3.88 & 2.32 & 5.57 & 5.33 & 7.86 & 4.64 \\
FLUX.1 Kontext & 2.90 & 4.04 & 2.38 & 5.16 & 5.25 & 7.91 & 4.61 \\
\bottomrule
\end{tabular}
}
\end{table}

\begin{figure}[t]
  \centering
  \includegraphics[width=\linewidth]{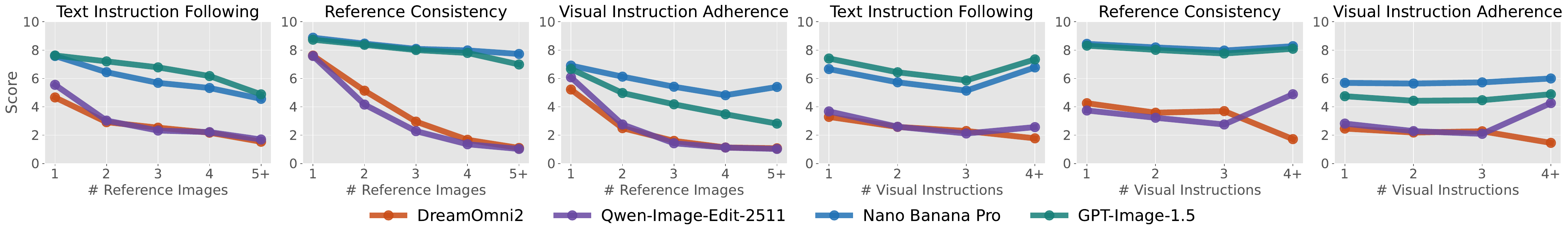}
  \caption{(\textbf{Left}) Scores versus the number of reference images for four representative generators. 
  More references degrade all three criteria and drive the open-weight models to near the floor in Reference Consistency and Visual Instruction Adherence, whereas the closed models largely retain Reference Consistency; among the closed models, Nano Banana Pro retains much of its Adherence while GPT-Image-1.5 drops sharply. 
  (\textbf{Right}) Scores versus the number of visual-instruction images.
  In contrast to the number of references, Adherence changes far less with more visual instructions and depends mostly on the generator itself, while additional visual instructions mainly lower Text Instruction Following.}
  \label{fig:num_refs_vis}
\end{figure}

\begin{figure}[t]
    \centering
    % \vspace{-0.1in}
    \begin{minipage}{0.60\linewidth}
        \centering
        \includegraphics[width=\linewidth]{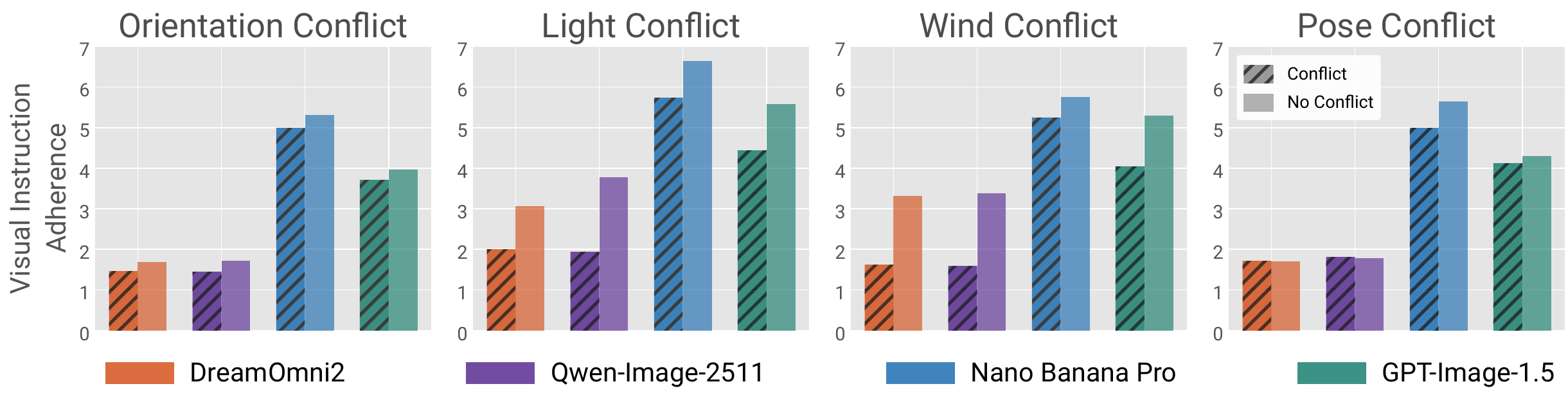}
    \end{minipage}\hfill
    \begin{minipage}{0.375\linewidth}
        \centering
        \includegraphics[width=\linewidth]{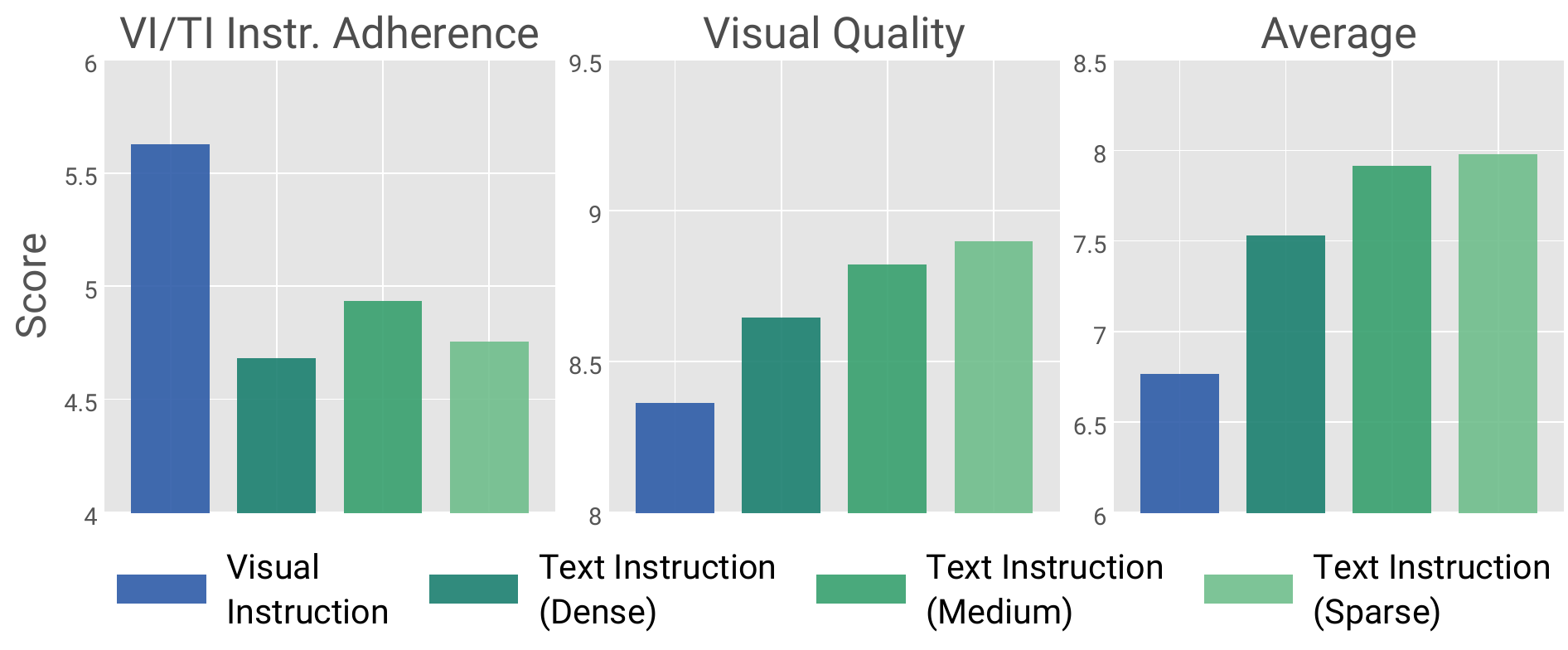}
    \end{minipage}
    % \vspace{-0.1in}
    \caption{(\textbf{Left}) Visual Instruction Adherence on conflict and no-conflict subsets for Orientation, Light, Wind, and Pose. Conflict generally reduces adherence, with particularly pronounced gaps for Light and Wind; for Pose, the reduction is visible only for generators with non-trivial adherence.
    (\textbf{Right}) Comparison between visual instructions (VIs) and text-converted instructions at different levels of granularity for Nano Banana Pro.
    VI and TI Dense approximately match information content and primarily differ in modality; TI Medium and TI Sparse progressively remove information. 
    All conditions are evaluated against the original VI, so the comparison among text variants measures recovery of the original constraint under increasing textual abstraction.
    }
    \label{fig:conflict_and_vi_vs_text}
    % \vspace{-0.15in}
\end{figure}

\subsection{Effect of the Number of References and Visual Instructions}
\label{sec:exp_n}

\autoref{fig:num_refs_vis} plots scores against the number of references and visual-instruction images for four representative generators (full results in Appendix~\ref{app:num_refs_vis}).
Adding references degrades all three criteria.
With a single reference, open-weight models trail closed ones only modestly in Reference Consistency and Visual Instruction Adherence, but with five or more references both fall to near the floor of the scale, whereas closed models largely retain their Reference Consistency.
The large open/closed gap in these criteria in \autoref{tab:per_vi} thus stems mainly from multi-reference tasks.
Closed models also diverge: Nano Banana Pro retains much of its Adherence, whereas GPT-Image-1.5 drops sharply, and this is where the adherence--artifact trade-off of Section~\ref{sec:exp_per_vi} emerges (Appendix~\ref{app:num_refs_vis}).

In contrast, from one to three visual instructions, Adherence remains nearly unchanged for both closed models and stays low for both open-weight models, whereas the same increase in references lowers it substantially.
Adherence thus reflects each generator's ability to interpret visual instructions more than their number; additional ones instead mainly lower Text Instruction Following.

\subsection{Reference--Visual-Instruction Conflict}
\label{sec:exp_conflict}

\autoref{fig:conflict_and_vi_vs_text} (\textbf{Left}) compares Visual Instruction Adherence between conflict and no-conflict subsets for each instruction type. 
Across the representative generators, the conflict subsets generally show lower adherence, although the gap varies by instruction type. 
The gap is largest for Light and Wind and smaller for Orientation. 
For Pose, the reduction appears in the generators that attain non-trivial adherence in the first place (Nano Banana Pro and GPT-Image-1.5), whereas the open-weight models already score near the floor on pose tasks regardless of conflict, leaving little room for a further drop. 
These results suggest that salient reference attributes may make it harder to follow visual instructions targeting the same attribute. 
Qualitative examples in \autoref{fig:qualitative} (\textbf{Right}) illustrate cases where models retain pose or lighting cues from the reference rather than fully following the target visual instruction. 
Further per-generator results and conflict-subset statistics are provided in Appendix~\ref{sec:further_results_vi_conflict}.

\subsection{How Should Visual Constraints Be Specified?}
\label{sec:exp_vi_vs_text}

Visual constraints can vary in representation and level of detail, which may affect how well models follow them.
We first isolate representation modality by comparing the original visual instruction (VI) with TI Dense, which verbalizes nearly all information encoded by the VI. 
For Nano Banana Pro, which shows strong VI-adherence in VIF-Bench, the original VI achieves higher adherence than its dense textual counterpart (\autoref{fig:conflict_and_vi_vs_text}; \textbf{Right}). 
This suggests that, when a model can reliably interpret visual instructions, presenting precise spatial or geometric constraints can be more effective than verbalizing the same information. 
In contrast, GPT-Image-1.5 benefits from text conversion, indicating that the preferred modality depends on the model's VI-understanding capability (Appendix~\ref{sec:further_results_vi_ti}).

We next consider a practical textual interface, where users may not verbalize every coordinate, angle, or joint configuration in a VI. 
We progressively abstract TI Dense into TI Medium and TI Sparse. 
These variants intentionally contain different amounts of information, and we evaluate all outputs against the original VI as an oracle specification. 
The score measures how well each textual representation recovers the original constraint, rather than how closely it adheres to the provided text. 
TI Medium recovers the original constraint better than TI Dense despite omitting fine-grained values, suggesting that reducing verbal complexity can offset some information loss. 
With TI Sparse, however, further simplification removes information needed to convey the original intent. 
Visual instructions thus offer a compact way to convey precise constraints without requiring exhaustive verbalization, while textual instructions require a balance between detail and abstraction.

\subsection{Agentic Multi-step Generation}
\label{sec:exp_multiturn}

We investigate whether VIF-Bench tasks can be solved more reliably by decomposing them into simpler sub-tasks.
GPT-5 plans $2$--$4$ sub-tasks and assigns a subset of the reference images to each; at every step, the generator receives the previous output, the sub-task instruction, and the newly assigned references, and its output is passed on.
The final image is evaluated against the original task.
This pipeline raises Nano Banana Pro's average score from $6.96$ to $7.40$ but slightly lowers GPT-Image-1.5's, from $7.60$ to $7.53$.
For both generators, however, the per-criterion changes follow the adherence--artifact pattern of Section~\ref{sec:exp_per_vi}: Visual Instruction Cleanliness rises (Nano Banana Pro: $6.27$ to $8.37$; GPT-Image-1.5: $9.39$ to $9.80$) while Visual Instruction Adherence falls (Nano Banana Pro: $5.67$ to $5.09$; GPT-Image-1.5: $4.57$ to $4.43$), indicating that the tension between the two objectives also holds within a single generator and that decomposition moves along the trade-off rather than escaping it.
Splitting the references across steps further reduces Reference Consistency by weakening subject grounding within each step.
The benefit of agentic decomposition therefore depends on the generator and the criterion.

\subsection{Reliability of VLM Judges}
\label{sec:exp_judge}

\begin{wraptable}{r}{0.42\textwidth}
    \centering
    \vspace{-0.15in}
    \caption{Correlation between human and VLM judges on average scores for the 168-image human-evaluation subset.
    Qwen3-VL-32B is included as an open-source judge.}
    \label{tab:judge_human}
    % \vspace{-0.12in}
    \scalebox{0.92}{
    \begin{tabular}{lcc}
        \toprule
        \textbf{Judge}
        & \textbf{Pearson $r$}
        & \textbf{Spearman $r$} \\
        \midrule
        GPT-5      & 0.78 & 0.75 \\
        Gemini 2.5 & 0.74 & 0.71 \\
        Qwen3-VL   & 0.73 & 0.70 \\
        \midrule
        Human      & 0.80 & 0.78 \\
        \bottomrule
    \end{tabular}
    }
    \vspace{-0.15in}
\end{wraptable}

To validate our VLM-based evaluation protocol, we measure Pearson's linear correlation coefficient and Spearman's rank-order correlation
coefficient between each VLM judge and human ratings on a uniformly randomly sampled subset of $168$ generated images.
We additionally report Human--Human agreement as a reference.
As shown in \autoref{tab:judge_human}, GPT-5 and Gemini 2.5 Flash both
show positive correlations with human evaluations across all criteria,
achieving correlation values of $0.78/0.75$ and $0.74/0.71$, respectively, for the Average Score.
Full results are shown in Appendix~\ref{sec:human_correlation_appendix}.

\section{Conclusion}
\label{sec:conclusion}

VIF-Bench enables systematic evaluation of multi-reference generation under heterogeneous visual constraints, including potential reference--VI conflicts and controlled VI––TI comparisons at different levels of specificity. 
Our evaluation reveals three findings: current models exhibit an adherence–artifact tension when following visual instructions; visual instruction adherence tends to be lower when references carry a salient state of the controlled attribute; and the best representation of a visual constraint depends on the model’s VI capability, with direct VIs benefiting strong VI-following models and intermediate textual specificity outperforming exhaustive verbalization. 
Together, these results highlight both the remaining limitations of multimodal generators and the importance of how visual constraints are represented.

% \clearpage
\subsection*{AI use statement}

In this work, we used generative AI tools to create synthetic datasets and implement methods.
In particular, generative models are integral components of the VIF-Bench pipeline described in the paper: Nano Banana and GPT-Image-1 generate part of the synthetic reference-image pool; GPT-5 proposes visual instructions (bounding-box layouts, 3D orientations, and wind/light arrow directions) and pre-labels reference attributes; and GPT-5, Gemini~2.5~Flash, and Qwen3-VL-32B serve as VLM judges in the evaluation protocol.
We have not used generative AI tools to develop theoretical models or conceptual frameworks, to propose or refine hypotheses, to design research methodology or experiments, or to interpret results, and the remaining required-disclosure tasks (formulating mathematical claims, providing critical ingredients for proving mathematical claims, and assisting in the writing of proofs) are not applicable to this work.
Additionally, we used generative AI tools to create and modify scientific figures and images, create and edit software code, create research artifacts, and draft parts of the manuscript.
We have reviewed all AI-assisted work.
All AI-proposed tasks and labels underwent automatic consistency checks and were verified and corrected by human annotators; the VLM-based judging protocol was validated against human ratings on a 168-image subset, including an open-source judge; AI-assisted code was verified and tested for correctness by the authors; and all AI-drafted text and figures were reviewed and edited by the authors, with technical claims checked against the experimental results.
We take responsibility for the final content of this work, including text, claims, and artifacts produced with the aid of generative AI.

% (This section is \textbf{required} and does not count toward the page limit.)

% In this work, we used generative AI tools for [tasks with required disclosure].
% We have not used generative AI tools for [other tasks with required disclosure],
% and [the rest of the required disclosure tasks] are not applicable to this work.
% Additionally, we used generative AI tools for [tasks with recommended
% disclosure]. We have reviewed all AI-assisted work. [Elaborate. For example, “we
% checked LLM-generated research ideas for potential plagiarism through a manual
% literature survey”, “LLM-generated code was verified and tested for correctness
% by 2 authors”, etc.]. We take responsibility for the final content of this work,
% including text, claims or artifacts produced with the aid of generative AI.

% See the ICLR 2027 AI Policy for Authors for more details. This statement should
% not be more than 1 page.

\subsection*{Ethics statement}

VIF-Bench is a diagnostic benchmark intended to advance the controllability and reliability of multi-reference image generation by exposing failure modes that are invisible to generic image-quality metrics, with positive implications for applications such as advertising, virtual try-on, and content creation, where faithfulness to user-specified composition matters more than aesthetic polish. At the same time, like any benchmark in the image-generation space, VIF-Bench indirectly contributes to the broader ecosystem of generative imaging, which carries well-known misuse risks, including deepfakes, non-consensual intimate imagery, identity fraud, and visual misinformation that can manipulate public opinion or harass individuals. The very capabilities VIF-Bench is designed to evaluate---faithful placement of specified subjects, control over facing direction, and adherence to global lighting and force cues--- also make synthetic media more convincing and, therefore, more dangerous when misused. We will mitigate this risk by releasing the benchmark as an evaluation-only resource that excludes model weights, and new generative tooling, and by building it on references drawn from existing public datasets, so that no new identifiable persons are introduced.

% (This section is \textbf{recommended} and does not count toward the page limit.)

% If authors feel that their paper submission raises questions regarding the Code
% of Ethics, they are encouraged to include a paragraph of Ethics Statement (at
% the end of the main text before references) to address potential concerns where
% appropriate. Topics include, but are not limited to, studies that involve human
% subjects, practices to data set releases, potentially harmful insights,
% methodologies and applications, potential conflict of interest and sponsorship,
% discrimination/bias/fairness concerns, privacy and security issues, legal
% compliance, and research integrity issues (e.g., IRB, documentation, research
% ethics). This statement should not be more than 1 page.

\subsection*{Reproducibility statement}

All results in this paper are collected using stable-version API endpoints for the VLM judges.
Additionally, we introduce an open-source judge (Qwen3-VL-32B) to ensure VIF-Bench remains evaluable even without closed-source API access.
We released the code (\url{https://github.com/shim0114/VIF-Bench}) and the benchmark (\url{https://huggingface.co/datasets/shim0114/VIF-Bench}).
% Finally, we anonymously released the code (\url{https://anonymous.4open.science/r/VIF-Bench-C213}).
% We will release our the benchmark.
%(\url{https://drive.google.com/drive/folders/1iCP0-W9jxXzSM8__Oyo2GiPao9FZ0oaG}).

% (This section is \textbf{recommended} and does not count toward the page limit.)

% It is important that the work published in ICLR is reproducible. Authors are
% strongly encouraged to include a paragraph-long Reproducibility Statement at the
% end of the main text (before references) to discuss the efforts that have been
% made to ensure reproducibility. This paragraph should not itself describe
% details needed for reproducing the results, but rather reference the parts of
% the main paper, appendix, and supplemental materials that will help with
% reproducibility. For example, for novel models or algorithms, a link to an
% anonymous downloadable source code can be submitted as supplementary materials;
% for theoretical results, clear explanations of any assumptions and a complete
% proof of the claims can be included in the appendix; for any datasets used in
% the experiments, a complete description of the data processing steps can be
% provided in the supplementary materials. Each of the above are examples of
% things that can be referenced in the reproducibility statement.

\subsubsection*{Acknowledgements}
We appreciate the funding support from Google Japan. MS was supported by JSPS KAKENHI Grant Number JP23H04974.

\bibliography{iclr2027_conference}
\bibliographystyle{iclr2027_conference}

\clearpage
\appendix
\section*{Appendix}

\section{Results of Qwen3-VL Judge}
\label{sec:qwen_results}

In Section~\ref{sec:experiments}, we report evaluation results based on the average scores assigned by Gemini 2.5 and GPT-5. 
As shown in Section~\ref{sec:exp_judge}, Qwen3-VL-32B-Instruct~\citep{bai2025qwen3vl} exhibits a high level of agreement with human evaluations, with correlations comparable to those of GPT-5~\citep{openai2025gpt5} and Gemini 2.5~\citep{geminiteam2023gemini} (see \autoref{tab:judge_human}). 
While we evaluate the closed-source judges using fixed API versions, their continued availability and exact reproducibility may still depend on external API access. 
We therefore additionally evaluate all outputs with Qwen3-VL-32B-Instruct, a fixed-version open-weight model, to provide a fully reproducible evaluation setting. This also makes the benchmark more accessible to researchers without access to proprietary VLM APIs.

\autoref{tab:qwen_judge} reports the scores obtained using Qwen3-VL-32B-Instruct as the sole evaluator, following the same scoring procedure described in Section~\ref{sec:evaluation}.
The generator ranking is preserved except for a swap between DreamOmni2 and FLUX.1 Kontext, which differ by less than 0.1 under every judge.
The adherence--artifact trade-off is also reproduced within each generator.
With multi-step decomposition, Visual Instruction Cleanliness rises (Nano Banana Pro: from 6.60 to 8.42; GPT-Image-1.5: from 9.48 to 9.85) while Visual Instruction Adherence falls (Nano Banana Pro: from 7.74 to 7.08; GPT-Image-1.5: from 7.90 to 7.19).
Likewise, the update from GPT-Image-1 to GPT-Image-1.5 raises Adherence from 7.40 to 7.90 while lowering Cleanliness from 9.81 to 9.48.
Qwen3-VL is more lenient on Adherence and compresses the four closed models into a 0.5-point band (7.40--7.90), so it does not resolve the cross-family gap on which GPT-5 and Gemini independently agree (see \autoref{tab:judge_specific_scores}); among the three judges, only GPT-5 reaches human--human agreement on this criterion (0.76/0.72 vs.\ 0.75/0.74; \autoref{tab:judge_human_full}).
These results support Qwen3-VL-32B-Instruct as a reproducible judge for overall comparison and indicate that our main conclusions do not hinge on the judge choice. We expect this fixed, open-weight judge to provide a useful baseline for future evaluations on our benchmark.

\begin{table}[ht]
\centering
\small
\caption{Overall VIF-Bench scores for each generator under the six evaluation criteria, evaluated by Qwen3-VL. 
We additionally report results for the agentic multi-step variants of GPT-Image-1.5 and Nano Banana Pro.}
% \vspace{-0.1in}
\label{tab:qwen_judge}
\scalebox{0.79}{
\begin{tabular}{l c c c c c c c}
\toprule
\textbf{Model} &
\makecell{\textbf{Text Instruction}\\\textbf{Following}} &
\makecell{\textbf{Reference}\\\textbf{Consistency}} &
\makecell{\textbf{Visual Instruction}\\\textbf{Adherence}} &
\makecell{\textbf{Visual Instruction}\\\textbf{Cleanliness}} &
\makecell{\textbf{Scene}\\\textbf{Coherence}} &
\makecell{\textbf{Visual}\\\textbf{Quality}} &
\textbf{Avg.} \\
\midrule
GPT-Image-1.5 & \textbf{8.27} & 9.34 & \textbf{7.90} & 9.48 & 8.87 & \textbf{9.74} & \textbf{8.93} \\
~ + multi-step & 7.88 & 8.75 & 7.19 & \textbf{9.85} & 8.85 & 9.69 & 8.70 \\
\midrule
Nano Banana Pro & 7.34 & 9.30 & 7.74 & 6.60 & 7.69 & 9.46 & 8.02 \\
~ + multi-step & 7.58 & 8.51 & 7.08 & 8.42 & 8.43 & 9.55 & 8.26 \\
\midrule
GPT-Image-1 & 8.01 & 9.27 & 7.40 & 9.81 & \textbf{8.95} & 9.72 & 8.86 \\
Nano Banana & 7.62 & \textbf{9.48} & 7.76 & 7.46 & 7.93 & 9.52 & 8.29 \\
Qwen-Image-2511 & 4.21 & 4.93 & 3.02 & 7.65 & 5.08 & 8.75 & 5.61 \\
Qwen-Image-2509 & 3.21 & 3.64 & 2.59 & 6.76 & 3.37 & 5.54 & 4.19 \\
DreamOmni2 & 4.36 & 5.45 & 3.42 & 5.72 & 4.55 & 8.61 & 5.35 \\
FLUX.1 Kontext & 4.57 & 5.74 & 3.63 & 5.30 & 4.65 & 8.73 & 5.44 \\
\bottomrule
\end{tabular}
}
\end{table}

\clearpage
\section{Implementation Details}
\label{sec:impl}

\paragraph{Code and Benchmark.}
The code is released at
\url{https://github.com/shim0114/VIF-Bench}, and the benchmark is released at \url{https://huggingface.co/datasets/shim0114/VIF-Bench}.
% \url{https://huggingface.co/datasets/shim0114/VIF-Bench}.
% \url{https://anonymous.4open.science/r/VIF-Bench-C213}.

% \paragraph{Benchmark.}
% We will release our benchmark.
% We released our benchmark anonymously at \url{https://drive.google.com/drive/folders/1iCP0-W9jxXzSM8__Oyo2GiPao9FZ0oaG}.

\paragraph{API versions.}
We used the following API endpoints. 
For the VLM judges, we fixed the endpoint versions to ensure reproducibility.
\begin{itemize}[leftmargin=0.5cm,topsep=0pt,itemsep=0pt]
\item For image generation models: \texttt{gemini-3-pro-image-preview}, \texttt{gemini-2.5-flash-image}, \texttt{gpt-image-1.5}, \texttt{gpt-image-1}.
\item For VLM evaluation: \texttt{gemini-2.5-flash}, \texttt{gpt-5-2025-08-07}.
\end{itemize}
Each task is generated with output size forced to $1024\!\times\!1024$.

\paragraph{Cost.}
A full judging pass over the $1{,}241$-task evaluated set with two judges for a single generator costs approximately \$$47$ end-to-end. The breakdown by judge is: GPT-5 $\sim$\$$35$ in total, and Gemini 2.5 Flash $\sim$\$$12$ in total.

% \clearpage
\section{Details of Task Construction}
\label{sec:appendix-task-construction}

\subsection{LAION-5B Image Filtering}
\label{sec:appendix-laion-filtering}

For reference images sourced from LAION-5B~\citep{schuhmann2022laion}, we applied the filtering procedure below.
First, to remove images that are unsuitable as references in terms of composition (foreground subject too small, or visually unclear), we ran YOLOv12~\citep{tian2025yolov12} object detection on every image, and then performed semantic segmentation with SAM~\citep{ravi2025sam} conditioned on the detected bounding boxes. We discarded an image if it satisfied any of the following: (i) the area of the bounding box covers less than 2\% of the entire image, (ii) the segmented region inside the bounding box covers less than 30\% of the box, or (iii) the CLIP similarity~\citep{radford2021clip} between the YOLO-predicted class name and the bounding-box region is below 20. On the other hand, we retained images with no detected objects, since we judged them useful as references for backgrounds or style-level content. After this stage, about 48\% of the original images remained.
Next, to remove images inappropriate as references (unsafe content, charts, screenshots of system messages, etc.), we combined automatic screening with Gemini~\citep{geminiteam2023gemini} and human review. For inappropriate content, we specifically targeted hate, harassment, violence, self-harm, sexual content, nudity, shocking content, illegal activity, and other distressing material, ultimately excluding about 3\% of the remaining images. We also manually verified and removed near-duplicate synthetic images.

\subsection{Language Coverage of Text References}
\label{sec:appendix-text-languages}

For the main text reference, we follow the design of MultiBanana~\citep{oshima2026multibanana} and cover three languages: English, Chinese, and Japanese. This lets us diversify not only typography but also language itself, both in tasks where text is the primary reference and in tasks where typography serves as a secondary reference.

\subsection{Visual Instruction Construction}
\label{sec:appendix-vi-construction}

Here we describe, for each visual instruction in our benchmark, the construction procedure and design choices we made to make the visual instructions easy to read as control signals and prevent them from leaking into the final generated image.

\noindent\textbf{Layout.}~~
For each main reference, we let GPT-5~\citep{openai2025gpt5} propose a 2D bounding box, which is then drawn as a colored rectangle on a 1024$\times$1024 canvas. In the proposal, we instruct the model to avoid collage-like or grid-like arrangements, prefer overlap, depth cues, and a common ground plane, ensure that no single box covers the entire canvas, and keep a margin of at least 0.05 from the canvas edges, so that the final generated image reads as a single natural picture. This prevents the layout instruction from being interpreted as an instruction to ``tile'' images side by side, which would make the final output look like a collage or split panel.
We also require the proposed bounding boxes to be tied to the canonical image index of the input (\texttt{Image\_0}, \texttt{Image\_1}, \dots). This prevents the correspondence between subjects and visual instructions from drifting in tasks with multiple references.

\noindent\textbf{Orientation.}~~
For each main reference whose orientation is meaningful (i.e., person, animal, or object), we have GPT-5 propose a 3D facing direction in terms of two angles: yaw and pitch. Here, yaw $= 0^\circ$ corresponds to the viewer/camera direction, $+90^\circ$ to screen-right, $-90^\circ$ to screen-left, and $180^\circ$ to the back, while positive pitch denotes upward and negative pitch downward. Based on these angles, we render a single 3D square pyramid (black background, gray fill, yellow-green outline) for each main reference. The pyramid's square base corresponds to the front of the subject, and the apex to the back.
In the proposal, we impose the visibility constraints $|yaw| < 60^\circ$, $|pitch| < 60^\circ$, and $|yaw| + |pitch| \leq 75^\circ$, and additionally forbid near-frontal angles ($|yaw| < 15^\circ$ and $|pitch| < 15^\circ$). This excludes angles where the pyramid collapses visually and the orientation becomes unreadable, and restricts the visual instruction to a range that remains interpretable.
We also make the role separation explicit: when a layout is available, the apex anchor uses the center of the layout box, and when no layout is available, the anchor position is purely for rendering and does not encode placement. This keeps orientation limited to facing direction, separate from layout's placement role.

\noindent\textbf{Wind and light arrow.}~~
We let GPT-5 propose, in normalized canvas coordinates, the start and end points of a 2D directed arrow. Wind arrows are drawn in cyan and light arrows in yellow; when both are present, they are drawn together on a single arrow image. The semantics are: for wind, the start denotes the wind source, and the end denotes the direction the wind blows toward; for light, the start denotes the light source, and the end denotes where the light falls.
In the proposal, we constrain the arrow length to be $\geq 0.4$ and require it to stay $\geq 0.05$ inside the canvas edges. This is because arrows that are too short or squashed against the edge are hard to read as directional cues, and because wind and light arrows can coexist in the same image, whereas fixing color and tail/head semantics prevents confusion between the two.

\noindent\textbf{Pose.}~~
Unlike other visual instructions, pose is specified not by coordinates but by adding a pose reference image. Pose references come from the curated pose pairs in VIBE~\citep{zhang2026vibe-benchmark} and are attached only when a person main reference is present. When a task has multiple person mains, we assign a pose reference to each and explicitly name the target person in the textual instruction. This avoids ambiguity about which person main a pose reference should apply to.

\subsection{Conflict Detection}
\label{sec:appendix-conflict-algorithm}

Here we describe the algorithm we use to label the conflict tasks.
The procedure consists of three stages: pre-labeling on the reference pool side, per-task label matching, and evaluation of the conflict condition for each visual-instruction kind.
First, for every image in the reference pool, we use GPT-5 to pre-assign attribute labels needed for conflict detection, including the subject's facing direction (yaw and the reliability of its estimate), the distinctiveness of the lighting (e.g., whether there is a special, colored, or localized light source), the wind responsiveness (whether the image contains elements such as hair, fabric, or smoke whose appearance changes under wind), and the distinctiveness of the pose (whether the subject is in a non-neutral pose with a clearly readable action).
Next, for each task, we read off the configuration of main references and visual instructions from the layout information and the reference hierarchy, and we link each main-reference image to its corresponding entry in the reference pool via hash matching, thereby making the pre-assigned attributes available for each main reference in the task.
Finally, for each visual instruction type included in the task, we evaluate the conditions described below; if at least one of the orientation, light, wind, or pose conditions is met, we label the task as the corresponding conflict task.

\noindent\textbf{Orientation conflict.}~~
An orientation conflict is declared when the subject in the reference image has a clear facing direction, and that direction disagrees with the direction specified by the visual instruction.
Concretely, we only enter the judgment when the facing direction estimated from the reference image (denoted source) is reliable, and we then compare it with the direction specified by the pyramid (denoted target). We declare a conflict when the minimum angular difference between source and target is $\geq 60^\circ$, or when the two have opposite signs.
We include sign mismatch to capture left/right flips such as ``the reference subject faces left, but the visual instruction specifies right.'' This is needed because relying on the angular difference alone would miss cases such as yaw $30^\circ$ and yaw $-20^\circ$: the difference is only $50^\circ$, but the left/right facing direction has actually flipped, and the appearance changes substantially.

\noindent\textbf{Light conflict.}~~
Light conflict occurs when the reference image contains distinctive lighting that strongly shapes the scene's appearance and the task includes a light arrow.
By distinctive lighting, we mean lighting that is clearly different from ordinary daylight or uniform indoor illumination, such as nighttime artificial light, neon, moonlight, window-beam light, spotlights, strong backlight, colored light, or any case where the lighting itself dominates the overall impression of the image.

\noindent\textbf{Wind conflict.}~~
Wind conflict applies when the reference image contains elements whose appearance changes substantially with wind, and the task includes a wind arrow.
Such elements include hair, clothing, fur, feathers, fabric, paper, smoke, and flames, whose shapes or flows change with wind direction and strength.

\noindent\textbf{Pose conflict.}~~
A pose conflict is declared when a person in the reference image is in a clear and distinctive pose, and a pose reference is attached to that person.
By distinctive pose, we mean a pose that is not neutral, such as a standing or front-facing still pose, but one in which the subject's action or posture is clearly readable from the image. Note that pose reference images taken from VIBE are by construction intended to express explicit poses, and are therefore treated as distinctive poses by default.

\noindent\textbf{Layout.}~~
A layout specifies a spatial arrangement over the whole scene rather than an intrinsic attribute of any reference image, so it is not subject to conflict detection.

\subsection{Converting Visual Instructions into Text Instructions}
\label{sec:appendix-vi-text}

\begin{figure}[t]
    \centering
    \includegraphics[width=\linewidth]{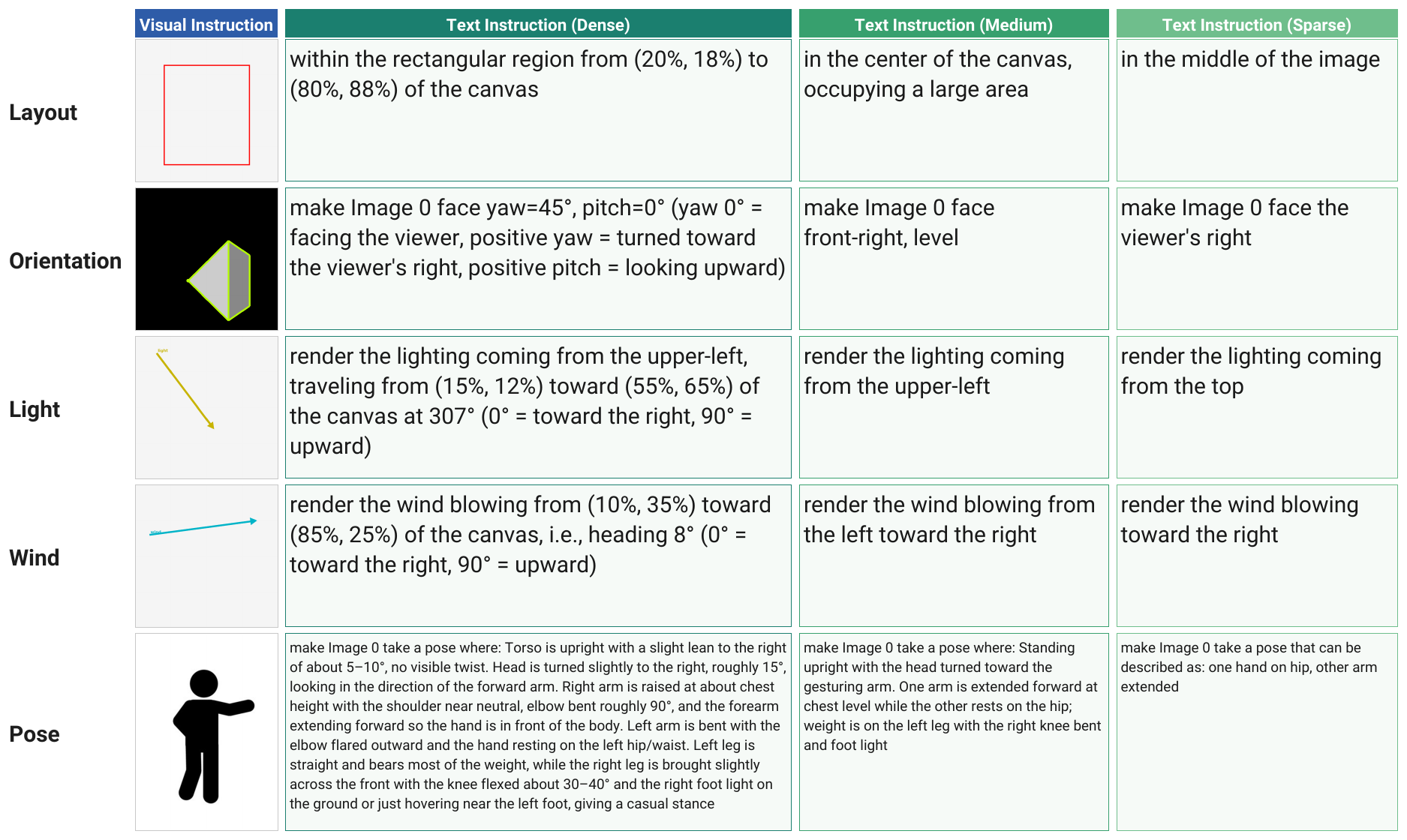}
    \caption{
    Examples of converting visual instructions (VIs) into text instructions (TIs) at three levels of granularity.
    From Dense to Sparse, information conveyed by the VI is progressively abstracted or removed.
    For example, Orientation is simplified from explicit yaw/pitch values to ``front-right, level'' and then to ``the viewer's right,'' while Pose is compressed from a detailed joint-level description to a short phrase capturing only its main configuration.
    }
    \label{fig:vi_ti_convert}
\end{figure}

Visual instructions (VIs), such as layout previews, 3D orientation pyramids, wind/light arrows, and pose references, provide a concise and intuitive way to specify spatial and directional constraints. Although the same constraints can in principle be expressed as text instructions (TIs), matching the precision of a VI may require explicitly verbalizing fine-grained information such as coordinates, viewing angles, and joint configurations. We therefore ask two questions: whether failures arise from the underlying constraint or from visual interpretation, and how the granularity of a textual description affects instruction following.

To this end, we replace each VI with one of three text variants: \textit{TI Dense}, \textit{TI Medium}, or \textit{TI Sparse}. \textit{TI Dense} preserves nearly all information in the original VI; \textit{TI Medium} replaces exact numerical values with coarser discrete descriptions; and \textit{TI Sparse} retains only the most salient spatial or directional attributes. Comparing VI with \textit{TI Dense} approximately isolates the effect of modality under matched information content, while the three TI variants reveal how performance changes as textual specificity is reduced.

All variants are constructed from the same 200-task subset, sampled with a fixed seed from tasks containing at least one non-layout VI and stratified to 50 tasks per subject count ($n{=}1$--$4$). The subset contains 96 tasks with an orientation instruction, 116 with a wind or light arrow, and 38 with a pose reference; every task contains a layout instruction. For each TI condition, we remove the corresponding VI images, re-index the remaining reference images, and remove the trailing clause asking the model to erase visual markings. Only the wording and granularity of the converted VI clauses differ across the three TI variants. Importantly, the judge is always given the \emph{original} task with the original VI images and instruction, rather than the rewritten TI, so the scores measure adherence to the original visual constraint rather than agreement with a coarsened textual description.

For layout, orientation, and arrow instructions, the underlying structured annotations---bounding boxes, yaw/pitch angles, and arrow endpoints--- let us construct the TI variants deterministically using templates. This gives precise control over the information retained at each granularity and avoids conversion errors from a language model. Pose references are available only as images, so for each unique pose image we use a single GPT-5 vision call to jointly produce the \textit{TI Dense}, \textit{TI Medium}, and \textit{TI Sparse} descriptions, keeping the three variants mutually consistent. \autoref{fig:vi_ti_convert} illustrates how each VI is progressively abstracted across the three text variants.

\paragraph{TI Dense.}
This variant preserves as much information as possible from the original VI. For layout, it specifies the exact bounding box using corner coordinates as percentages of the canvas (e.g., ``within the rectangular region from (13\%, 8\%) to (87\%, 92\%) of the canvas''). For orientation, it gives the exact yaw and pitch in degrees together with the sign convention. For arrows, it specifies the start and end points in canvas percentages and the heading angle; wind is described as blowing from the start point toward the end point, while light is described as arriving from the arrow's origin. For pose, it provides a three-to-five-sentence joint-level description covering the torso, head, arms, and legs, including approximate angles and heights. \textit{TI Dense} is thus the closest textual counterpart to the original VI, but requires information conveyed directly by the VI to be explicitly verbalized.

\paragraph{TI Medium.}
This variant replaces continuous or fine-grained quantities with a small number of discrete categories. For layout, it specifies the $3{\times}3$ grid cell containing the bounding-box center together with a three-level size label (small, medium, or large), while discarding the exact corners. Orientation is reduced to one of eight facing directions and three pitch levels. Arrow direction is quantized into eight compass directions, yielding descriptions such as ``blowing from left to right'' or ``coming from the upper-left.'' Pose is summarized in one or two sentences describing the torso, arms, and legs without numerical values. Compared with \textit{TI Dense}, this representation is easier to specify but omits exact positions and angles.

\paragraph{TI Sparse.}
This variant retains only coarse spatial or directional information. Layout specifies only the horizontal third of the image---left, middle, or right---without vertical position or size. Orientation is reduced to four directions with no pitch information, and arrow direction to four cardinal directions. Pose is represented by a short phrase describing its main configuration, such as ``one hand on hip, other arm extended.'' \textit{TI Sparse} substantially reduces textual complexity, but discards information such as box size, vertical position, fine-grained angles, and individual joint configurations.

\clearpage
\section{Further Statistics}
\label{sec:appendix-further-stats}

\begin{figure}[ht]
    \centering
    \includegraphics[width=\linewidth]{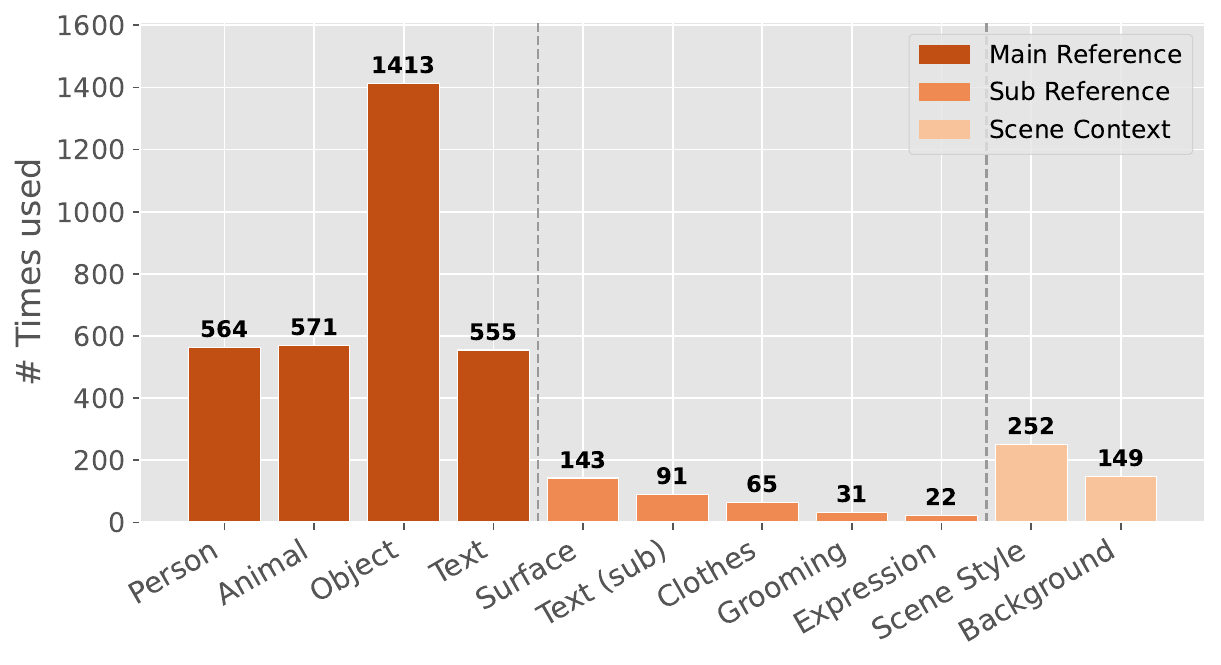}
    \caption{Reference adoption counts on the VIF-Bench evaluated set (1{,}241 tasks).
    Each bar reports how many times a reference type is used, grouped into
    Main Reference, Sub Reference, and Scene Context.}
    \label{fig:reference_adoption}
\end{figure}

\autoref{fig:reference_adoption} shows how often each reference type is adopted across the 1{,}241 tasks of our evaluated set, grouped into Main Reference, Sub Reference, and Scene Context and colored consistently with the reference-count distribution in \autoref{fig:task_stats}.
Main references (the primary subjects) are adopted 564 times for \textit{person}, 571 for \textit{animal}, 1{,}413 for \textit{object}, and 555 for \textit{text}. The object count is larger than the others because, unlike the single-form person, animal, and text categories, objects are further split by their plausible placement environment into \textit{versatile} (usable both indoors and outdoors), \textit{indoor-only}, and \textit{outdoor-only}; the object bar therefore aggregates three sub-types, breaking down into 512 versatile, 459 indoor-only, and 442 outdoor-only adoptions. Consequently, person, animal, and text remain at a comparable scale (roughly 550–570 adoptions each), while object—being an umbrella over three environment-conditioned variants—accumulates a proportionally larger total.
Sub-references specify attributes of a main subject and are therefore tied to a specific main-subject type: on the person side we adopt \textit{clothes} 65 times, \textit{grooming} (hairstyle and make-up) 31 times, and \textit{facial expression} 22 times; on the object side we adopt \textit{surface} (color and material) 143 times; and on the text side \textit{text} 91 times. Because each sub-reference type is only applicable to a compatible main subject—clothes, grooming, and facial expression apply to people, surface applies to objects, and text applies to text subjects—a sub-reference can be adopted only when the corresponding main type is present in the task. Consequently, the sub-reference counts are inherently uneven: they are upper-bounded by how often the associated main category appears (e.g., person for grooming, object for surface) and by whether a task chooses to specify that attribute. This subject-conditioned hierarchy also prevents semantically inconsistent task construction, such as assigning a facial-expression reference to an object.
Scene context, which specifies the overall appearance of the generated scene, is adopted 252 times for \textit{scene style} and 149 for \textit{background}.

\clearpage
\section{Comparison with Prior Works}
\label{sec:prior_work_comparison_appendix}

\autoref{tab:benchmarks} compares benchmarks by the maximum number of visual-instruction images provided within a single task. 
This image-level definition is distinct from the semantic number of visual operations encoded in those images. 
In VIBE~\citep{zhang2026vibe-benchmark}, multiple visual operations may be combined within a single annotated image in its multi-task setting; we therefore count it as one visual-instruction image at the image level, rather than treating it as semantically containing only a single instruction. 
MultiRef~\citep{chen2025multiref} supports multiple reference images, but each task uses at most a single visual-instruction image. 
DreamOmni3~\citep{xia2025dreamomni3} can involve up to two scribble-annotated images in a task, but these use a single scribble-based instruction modality rather than multiple heterogeneous visual-instruction types.
VIF-Bench instead requires models to jointly process multiple independent reference images together with multiple visual-instruction images, integrating heterogeneous constraints such as layout, orientation, pose, wind, and lighting.

\begin{figure*}[ht]
    \centering
    \includegraphics[width=0.9\linewidth]{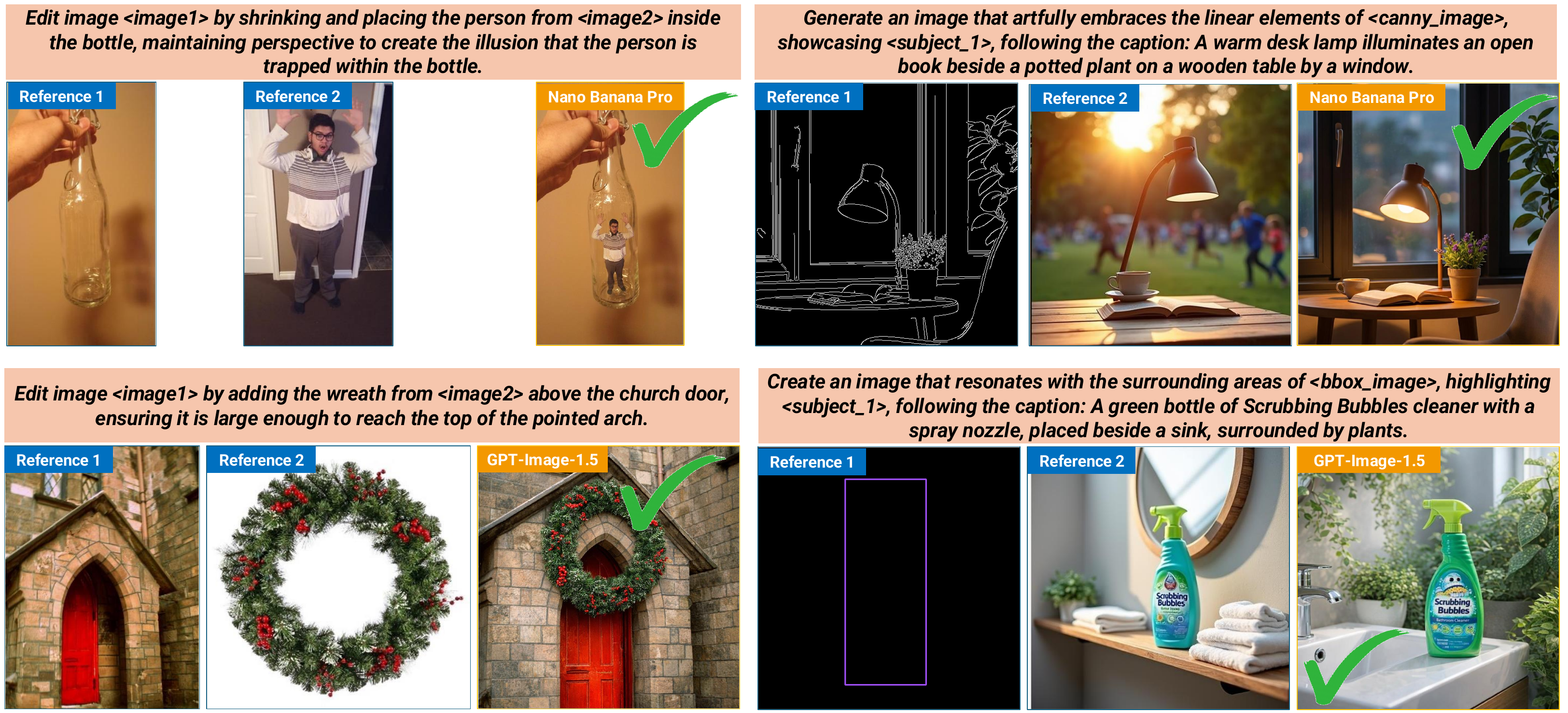}
    % \vspace{-0.2in}
    \caption{Example of MultiRef benchmark. These tasks are almost fully solvable by advanced models such as Nano Banana Pro and GPT-Image-1.5, and each task uses at most a single visual-instruction image.}
    \label{fig:qualitative_multiref}
\end{figure*}

Beyond comparison in \autoref{tab:benchmarks}, we also examine how recent state-of-the-art image generators perform on an existing multi-reference benchmark. 
We evaluate Nano Banana Pro and GPT-Image-1.5 on MultiRef~\citep{chen2025multiref}, which assesses heterogeneous visual-reference conditions using three Overall Assessment dimensions: Image Quality (IQ), Instruction Following (IF), and Source Fidelity (SF). 
As shown in \autoref{tab:multiref_sota}, GPT-Image-1.5 achieves an average score of 0.848, exceeding the 0.771 score of the ground-truth images under MultiRef's Overall Assessment protocol. 
This suggests that the current MultiRef evaluation provides relatively limited headroom for distinguishing the strongest recent generators.
These quantitative findings are echoed by qualitative inspection (\autoref{fig:qualitative_multiref}). 

\begin{table}[ht]
\centering
\small
\caption{
Comparison with prior methods on MultiRef~\citep{chen2025multiref}.
IQ, IF, and SF denote Image Quality, Instruction Following, and
Source Fidelity, respectively.
}
\label{tab:multiref_sota}
\scalebox{1.0}{
\begin{tabular}{lcccc}
\toprule
Model & IQ & IF & SF & Avg. \\
\midrule
Show-o~\citep{xie2024showo}            & 0.764 & 0.616 & 0.462 & 0.614 \\
OmniGen~\citep{xiao2024omnigen}           & 0.730 & 0.532 & 0.438 & 0.567 \\
ACE~\citep{han2025ace}               & 0.740 & 0.655 & 0.528 & 0.641 \\
ChatDiT~\citep{lhhuang2024chatdit}           & 0.811 & 0.713 & 0.574 & 0.699 \\
Claude + SD 2.1~\citep{rombach2022ldm}   & 0.812 & 0.726 & 0.572 & 0.703 \\
Claude + SD 3~\citep{esser2024sd3}     & 0.876 & 0.817 & 0.658 & 0.784 \\
Claude + SD 3.5~\citep{esser2024sd3}   & 0.913 & 0.853 & 0.691 & 0.819 \\
Gemini + SD 2.1~\citep{rombach2022ldm}   & 0.791 & 0.708 & 0.547 & 0.682 \\
Gemini + SD 3~\citep{esser2024sd3}     & 0.856 & 0.804 & 0.639 & 0.766 \\
Gemini + SD 3.5~\citep{esser2024sd3}   & 0.893 & 0.839 & 0.676 & 0.803 \\
Ground Truth      & 0.842 & 0.803 & 0.668 & 0.771 \\
\midrule
Nano Banana Pro~\citep{google2025nanobananapro}   & 0.800 & 0.743 & 0.660 & 0.734 \\
\textbf{GPT-Image-1.5}~\citep{openai2025gptimage1_5}
& \textbf{0.879}
& \textbf{0.856}
& \textbf{0.810}
& \textbf{0.848} \\
\bottomrule
\end{tabular}
}
\end{table}

\clearpage
\section{Further Results}

\subsection{Further Results for Reference--Visual-Instruction Conflict}
\label{sec:further_results_vi_conflict}

We provide a more detailed analysis of reference--visual-instruction conflict across all evaluated generators.
Among the 1,241 tasks in VIF-Bench, 438 contain at least one conflict between an attribute implied by a reference image and a visual instruction controlling the same attribute, while 803 contain no such conflict.
At the instruction-type level, conflicts occur in 106 of the 315 Orientation tasks (33.7\%), 172 of the 233 Light tasks (73.8\%), 153 of the 250 Wind tasks (61.2\%), and 82 of the 141 Pose tasks (58.2\%).
These categories are not mutually exclusive, since a single task may contain conflicts for multiple visual-instruction types.

\begin{figure*}[ht]
    \centering
    \includegraphics[width=\linewidth]{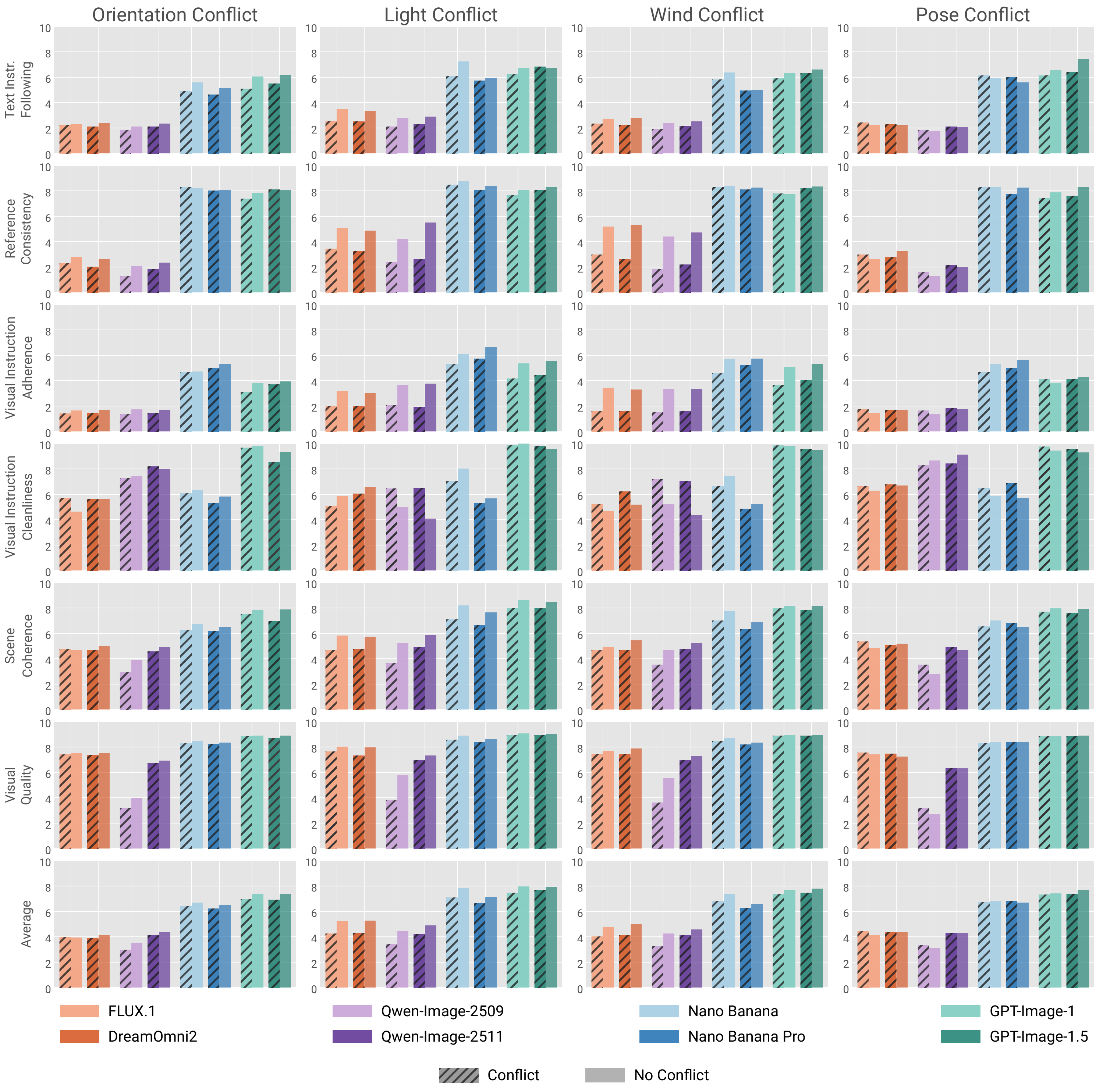}
    \caption{
    Full per-generator analysis of reference--visual-instruction conflict. For each visual-instruction type (columns), we compare conflict (hatched) and no-conflict subsets across all eight generators on the six evaluation criteria and their average (rows). On Visual Instruction Adherence (third row), conflict consistently reduces scores for Orientation, Light, and Wind, with substantially larger gaps for Light and Wind, whereas the effect for Pose is more model-dependent.
    }
    \label{fig:conflict_full}
\end{figure*}

Figure~\ref{fig:conflict_full} extends the analysis in Figure~\ref{fig:conflict_and_vi_vs_text} (\textbf{Left}) to all eight generators and all evaluation criteria. For each visual-instruction type, we split the corresponding tasks into conflict and no-conflict subsets and report the average score on each criterion for each generator. Below, we focus on Visual Instruction Adherence (third row), the criterion directly targeted by the conflict.
The full results show that the effect of conflict depends strongly on the controlled attribute.
For Orientation, adherence is lower on the conflict subset for every generator, although the gap is relatively modest.
The effect is substantially larger for Light and Wind: all eight generators obtain lower adherence on conflict tasks, indicating that salient lighting conditions or wind-responsive appearance already present in a reference can strongly interfere with the corresponding visual instruction.

The raw per-generator scores further show that this effect is not simply driven by a particular model family.
For Light, the conflict--no-conflict decrease ranges from approximately 0.78 to 1.85 points across the eight generators, while for Wind it ranges from approximately 0.51 to 1.86 points.
For Orientation, the decrease is smaller, ranging from approximately 0.08 to 0.68 points.
A simple average over the eight per-generator scores gives conflict versus no-conflict adherence of 2.77 versus 3.08 for Orientation, 3.46 versus 4.67 for Light, and 2.99 versus 4.43 for Wind.

Pose exhibits a different pattern.
The effect is more model-dependent: Nano Banana, Nano Banana Pro, and GPT-Image-1.5 show lower adherence on pose-conflict tasks, whereas several open-weight models show little difference or a small change in the opposite direction.
These open-weight models score near the floor on pose tasks in both subsets, so the absence of a gap likely reflects a floor effect rather than robustness to conflict.
Accordingly, averaging across all eight generators yields only a small aggregate difference for Pose (3.11 versus 3.17).
This suggests that reference--visual-instruction interference is particularly systematic for Light and Wind, while the effect of a conflicting reference pose depends more strongly on the generator.

\clearpage
\subsection{Further Results for the Number of References and Visual Instructions}
\label{app:num_refs_vis}

\begin{figure}[t]
    \centering
    % \vspace{-0.1in}
    \begin{minipage}{0.49\linewidth}
        \centering
        \includegraphics[width=\linewidth]{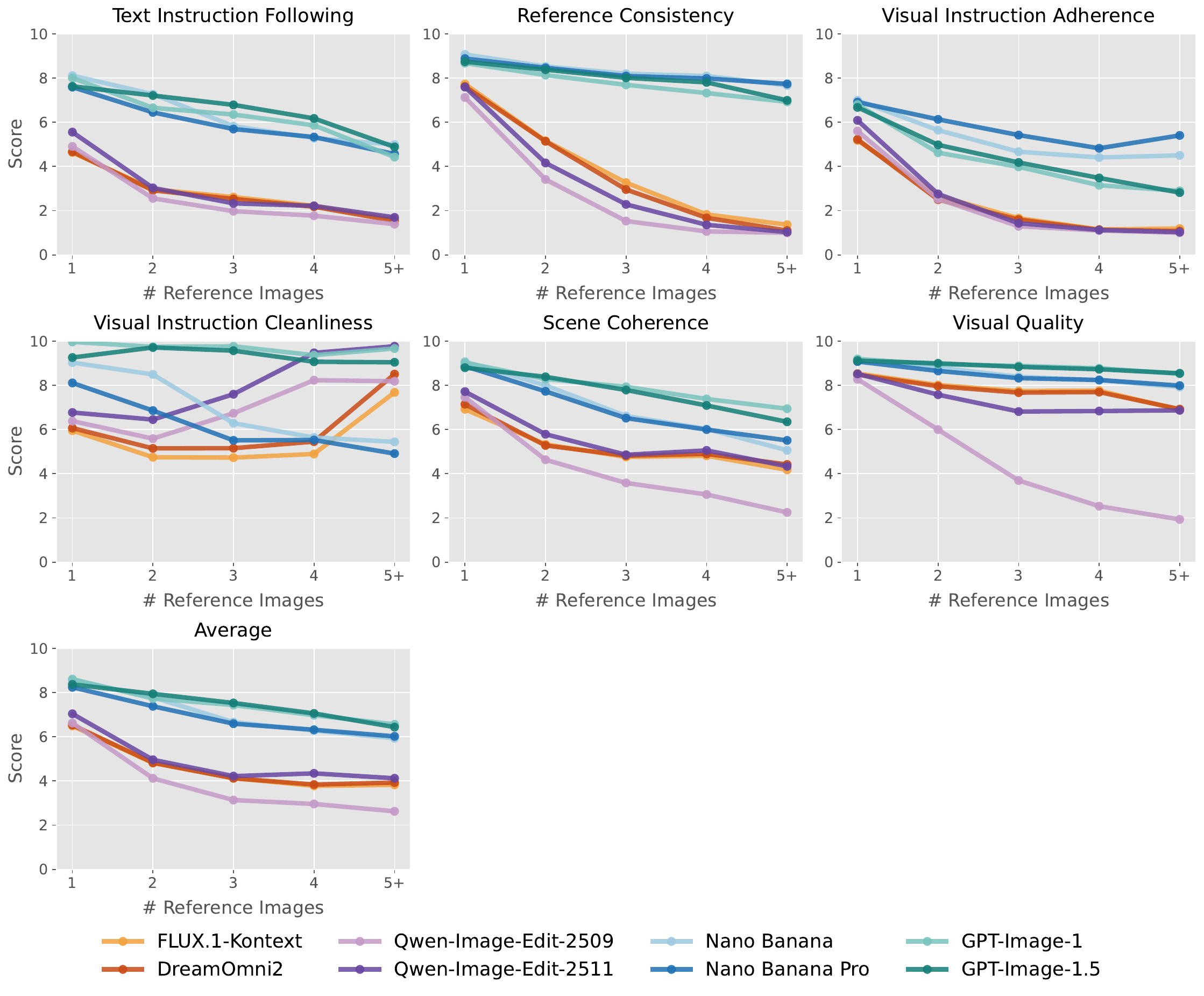}
    \end{minipage}\hfill
    \begin{minipage}{0.49\linewidth}
        \centering
        \includegraphics[width=\linewidth]{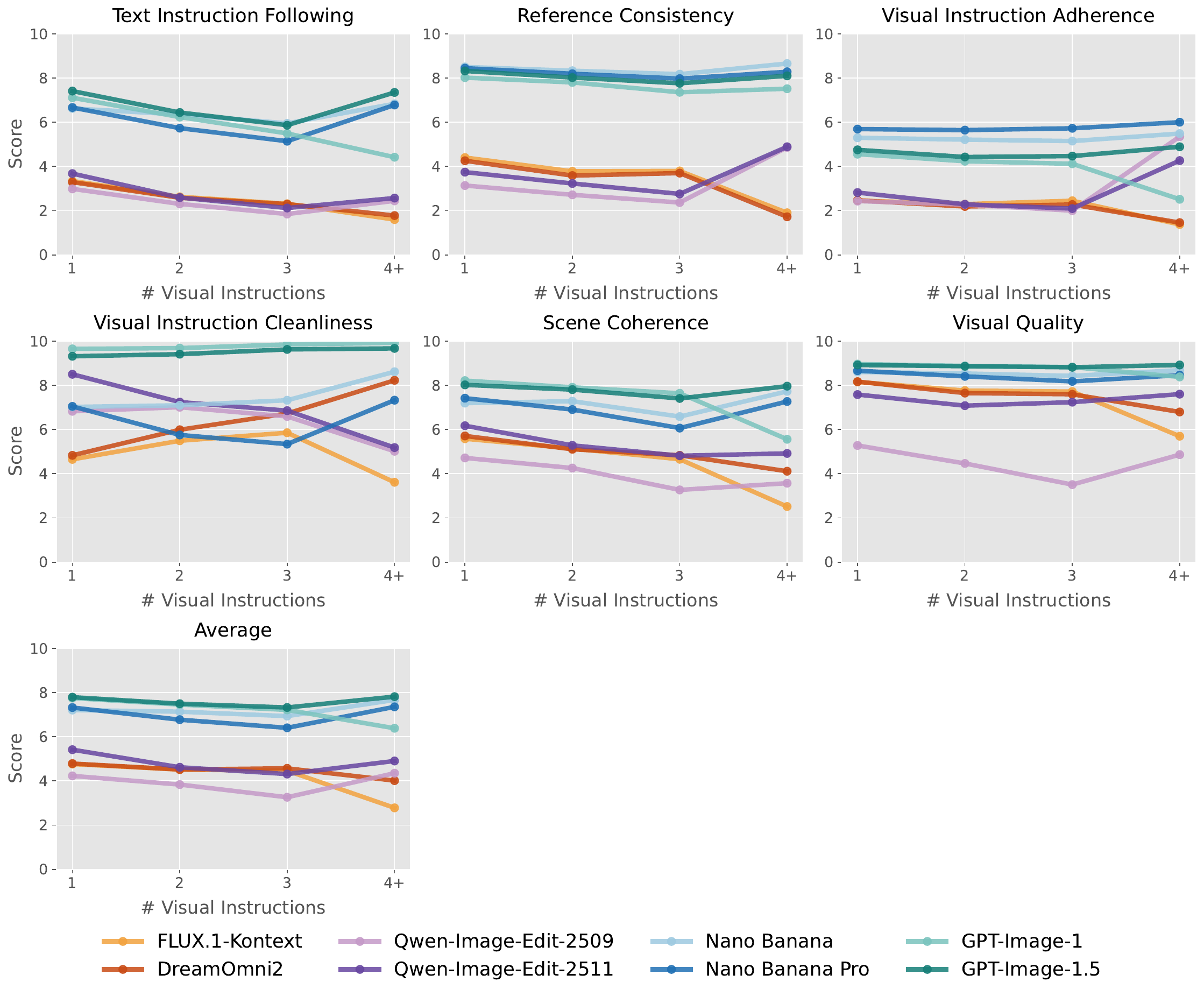}
    \end{minipage}
    % \vspace{-0.1in}
    \caption{(\textbf{Left}) Full results as a function of the number of reference images for all generators. We merge tasks with five or more reference images into the 5+ bin. For every generator, all criteria except Visual Instruction Cleanliness are lower with five or more references than with one. Cleanliness decreases for the Nano Banana models but increases for the open-weight generators, whose adherence approaches the floor.
    (\textbf{Right}) Full results as a function of the number of visual-instruction images for all generators. Tasks with four or more visual instructions are merged into the 4+ bin, which contains few tasks. The effect is weaker than that of the number of reference images.
    }
    \label{fig:num_refs_full}
    % \vspace{-0.15in}
\end{figure}

\autoref{fig:num_refs_full} extends \autoref{fig:num_refs_vis} to all generators and all criteria, together with their average.
As in Section~\ref{sec:exp_n}, tasks are binned by the number of reference images or visual-instruction images.

\paragraph{Number of reference images.}
Every generator scores lower with five or more references than with one on every criterion except Visual Instruction Cleanliness.
Reference Consistency separates the two model groups most clearly: the open-weight generators start relatively close to the closed ones with a single reference but fall to near the floor with five or more, whereas the closed generators decline only moderately.
Since the judge assigns the minimum Reference Consistency score whenever any subject is missing or replaced (Appendix~\ref{sec:prompts}), these near-floor scores suggest that, according to the judges, at least one referenced subject is missing or replaced in most of these outputs.
Visual Instruction Adherence for the open-weight generators likewise falls near the floor.
Text Instruction Following, in contrast, drops by a similar amount in both groups, so its open/closed gap remains roughly constant.
Visual Quality of the closed generators stays high in every bin, and Qwen-Image-Edit-2509 is the only generator whose Visual Quality collapses.
Its successor, Qwen-Image-Edit-2511, starts at a similar level but declines only moderately, so the Visual Quality gain from 2509 to 2511 in \autoref{tab:per_vi} arises largely from multi-reference tasks.
Scene Coherence declines for all generators, more steeply for the Nano Banana models than for the GPT-Image models.

\paragraph{Emergence of the adherence--artifact trade-off.}
Visual Instruction Cleanliness follows family-dependent trends that connect to the adherence--artifact trade-off in Section~\ref{sec:exp_per_vi}.
With a single reference, the closed generators are nearly tied in Visual Instruction Adherence, and all keep Cleanliness high.
With five or more references, they split into two groups: the Nano Banana models retain higher adherence but leave more instruction marks, whereas the GPT-Image models remain clean but lose much of their adherence.
The trade-off among closed models thus emerges as references accumulate.
For the open-weight generators, Cleanliness is instead higher with five or more references than with one.
Because this increase coincides with near-floor adherence, it suggests that their outputs neither follow nor reproduce the visual instructions, but rather than following them cleanly.

\paragraph{Number of visual instructions.}
Compared with references, the number of visual instructions has a weaker effect.
From one to three visual instructions, the Average score decreases for every generator, but always less than it does over the same increase in references; the two effects are closest for Nano Banana Pro and the GPT-Image models.
For every generator, Visual Instruction Adherence changes only slightly and far less than with references, even though the judge caps adherence by the worst violation among all visual instructions (Appendix~\ref{sec:prompts}).
Instead, Text Instruction Following, Reference Consistency, Scene Coherence, and Visual Quality decrease for every generator, typically most in Text Instruction Following and Scene Coherence.
Visual Instruction Cleanliness moves in both directions: it drops for Nano Banana Pro and Qwen-Image-Edit-2511 but rises for DreamOmni2 and FLUX.1 Kontext.
The 4+ bin contains few tasks and shows large model-dependent deviations in both directions; for example, the Average score drops for GPT-Image-1 but rises for GPT-Image-1.5.

\subsection{Further Results for Visual vs.\ Text-Converted Instructions}
\label{sec:further_results_vi_ti}

\begin{figure*}[ht]
    \centering
    \begin{minipage}[t]{\textwidth}
        \centering
        \includegraphics[width=\linewidth]{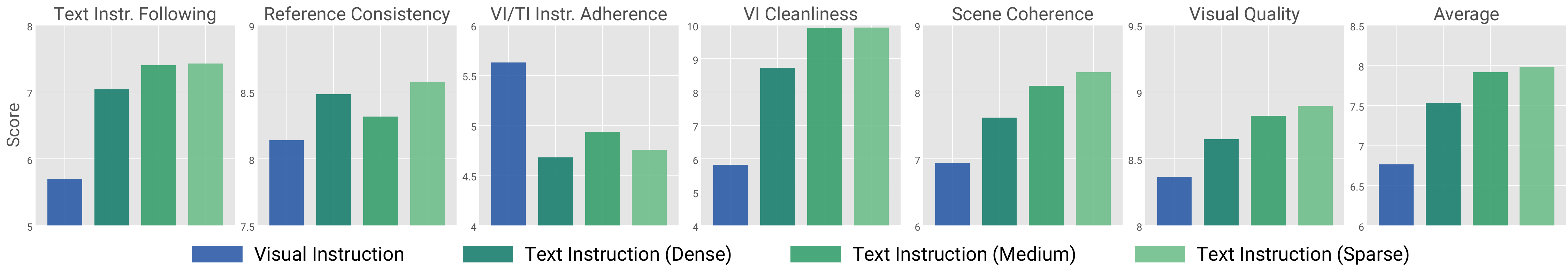}
        % \vspace{-0.08in}
        \small (a) Nano Banana Pro
    \end{minipage}
    % \hfill
    \begin{minipage}[t]{\textwidth}
        \centering
        \includegraphics[width=\linewidth]{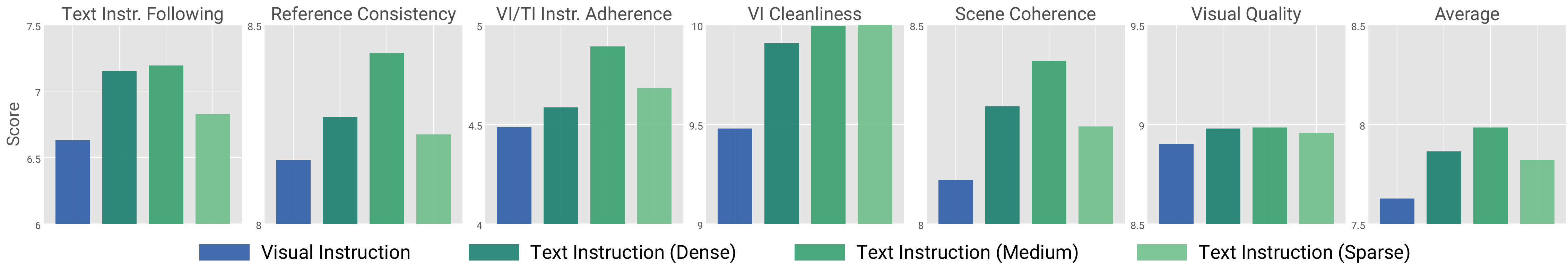}
        % \vspace{-0.08in}
        \small (b) GPT-Image-1.5
    \end{minipage}
    % \vspace{-0.05in}
    \caption{
    Full comparison of visual instructions (VIs) and text-converted instructions at three levels of granularity for Nano Banana Pro and GPT-Image-1.5.
    Nano Banana Pro achieves its highest VI/TI Instruction Adherence with the original VI, whereas GPT-Image-1.5 benefits from text conversion and performs best with TI Medium.
    Other criteria generally improve as the constraints are relaxed; in particular, Visual Instruction Cleanliness increases when VI images are removed.
    We evaluate all conditions against the original task and its original VIs.
    }
    \label{fig:further_vi_ti}
    % \vspace{-0.10in}
\end{figure*}

\autoref{fig:further_vi_ti} reports the full results for Nano Banana Pro and GPT-Image-1.5 across all evaluation criteria. Unlike Nano Banana Pro, whose Visual Instruction Adherence is highest with the original VI, GPT-Image-1.5 has relatively weak VI adherence and benefits from converting the same constraints into text. 
Even for GPT-Image-1.5, the most detailed \textit{TI Dense} is not optimal: \textit{TI Medium} achieves the highest VI/TI Instruction Adherence, suggesting that, for this model, moderately abstracted descriptions are easier to follow than either the visual instructions or exhaustive textual ones.
For the remaining metrics, performance generally improves as the instruction becomes less restrictive, reflecting greater freedom to preserve reference content and overall image quality. Visual Instruction Cleanliness also increases substantially when moving from VI to TI, as the visual instruction images---and hence the marks that could be reproduced in the output---are no longer provided. Importantly, we evaluate all generations using the VLM judges against the \emph{original task}, including its original VI images, rather than against the converted TI. We use the original VI as the oracle specification so that the comparison measures how well each textual representation recovers the constraint encoded by the VI, rather than how well the output matches a potentially coarsened textual instruction.

\clearpage
\subsection{Additional Qualitative Results}
\label{sec:additional_qualitative}

\begin{figure}[ht]
    \centering
    \includegraphics[width=\linewidth]{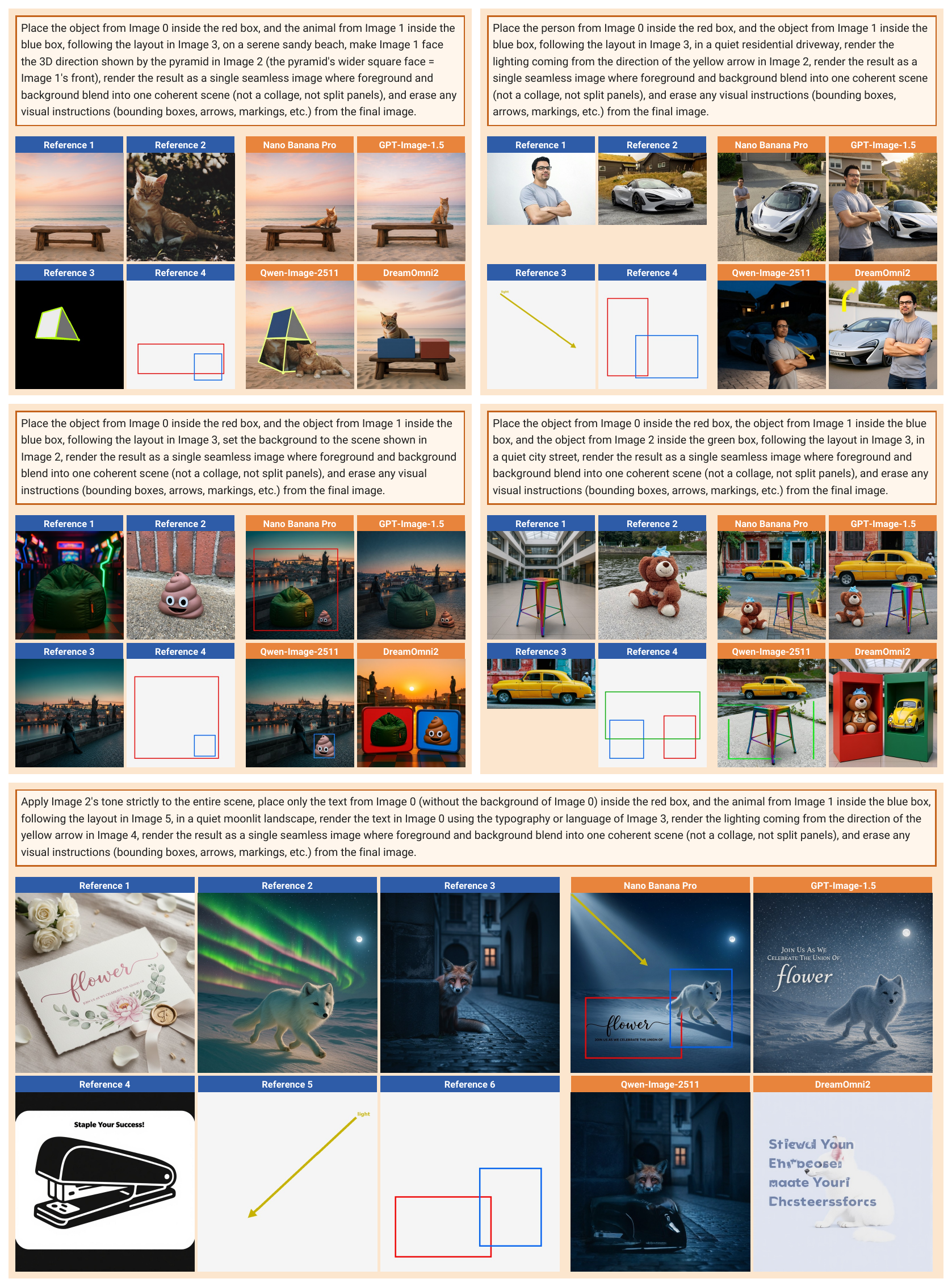}
    \caption{Qualitative example of VIF-Bench.}
    \label{fig:qualitative_appendix_1}
\end{figure}

\begin{figure}[t]
    \centering
    \includegraphics[width=0.95\linewidth]{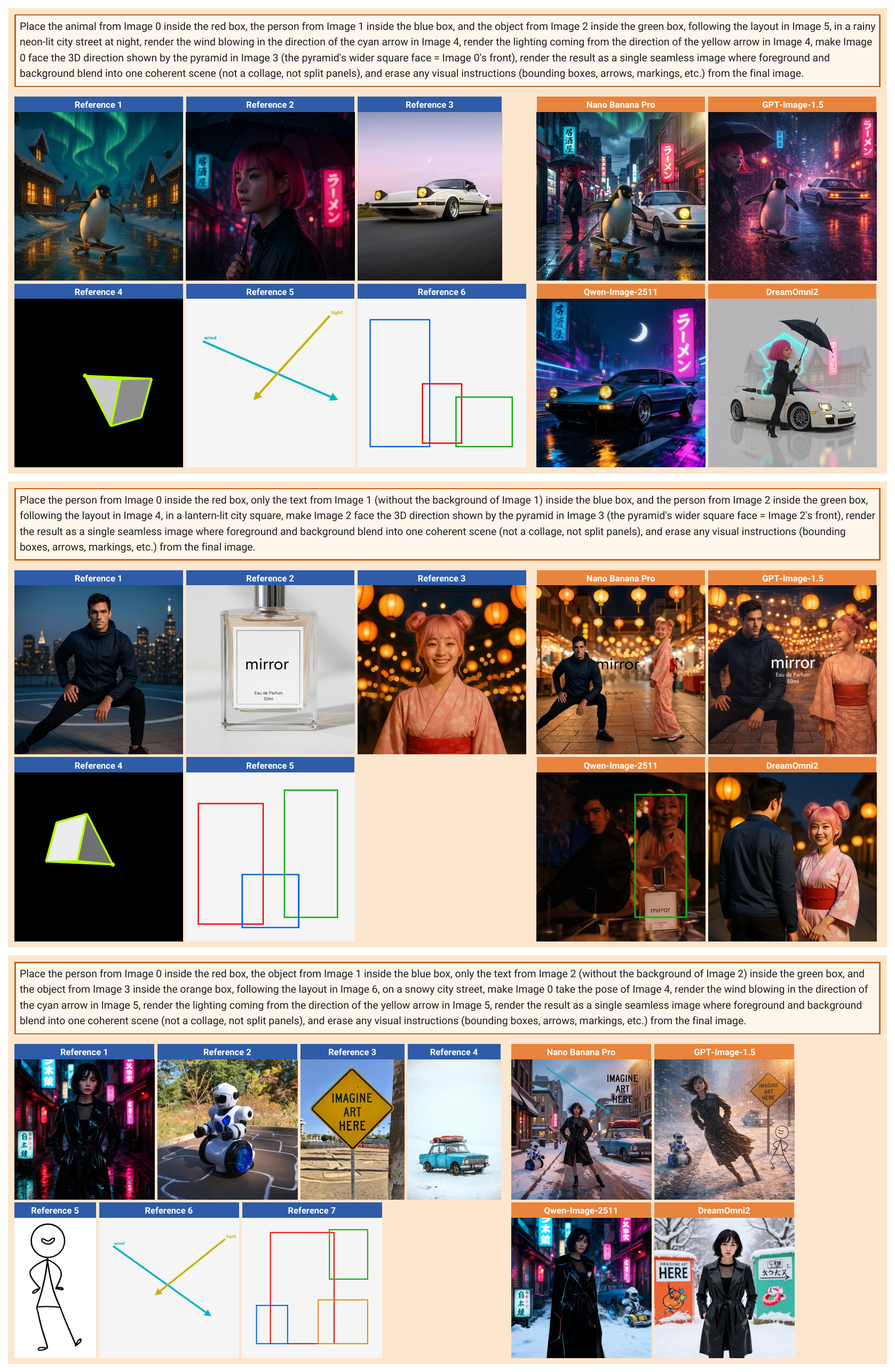}
    \caption{Qualitative example of VIF-Bench.}
    \label{fig:qualitative_appendix_2}
\end{figure}

\autoref{fig:qualitative_appendix_1} and \autoref{fig:qualitative_appendix_2} show additional qualitative examples of VIF-Bench.

\clearpage

\clearpage
\section{Further Discussion for Evaluation}
\label{sec:appendix_further_resutls}

\subsection{Robustness to Judge-Specific Preferences}
\label{app:judge_specific}

To examine whether the benchmark conclusions depend on preferences specific to a particular VLM judge, we report the evaluation results separately for GPT-5 and Gemini 2.5 Flash.
Each entry in \autoref{tab:judge_specific_scores} reports
\textbf{GPT-5 / Gemini}.
Although GPT-5 and Gemini assign somewhat different absolute scores on individual criteria, their Average Scores produce the same generator ranking.
This indicates that VIF-Bench's overall conclusions are robust to the choice between the two primary judges and are unlikely to reflect preferences specific to either model.

\begin{table}[ht]
\centering
\small
\caption{
Generator scores evaluated separately by GPT-5 and Gemini 2.5 Flash.
Each entry reports \textbf{GPT-5 / Gemini}.
}
\label{tab:judge_specific_scores}
\scalebox{0.88}{
\begin{tabular}{lccccccc}
\toprule
Model
& Text Instr.
& Ref. Consist.
& Vis. Adher.
& Vis. Clean.
& Scene
& Quality
& Avg. \\
\midrule
GPT-Image-1
& 6.39 / 6.63
& 8.08 / 7.59
& 4.16 / 4.54
& 9.74 / 9.63
& 7.62 / 8.37
& 8.95 / 8.86
& 7.49 / 7.60 \\

GPT-Image-1.5
& 6.70 / 6.89
& 8.45 / 7.79
& 4.22 / 4.91
& 9.41 / 9.37
& 7.51 / 8.16
& 8.99 / 8.77
& 7.55 / 7.65 \\

Nano Banana Pro
& 6.12 / 6.02
& 8.82 / 7.72
& 5.52 / 5.82
& 6.30 / 6.24
& 7.17 / 6.84
& 8.83 / 8.13
& 7.13 / 6.80 \\

Nano Banana
& 6.42 / 6.38
& 8.84 / 7.93
& 4.93 / 5.54
& 7.13 / 7.05
& 7.11 / 7.15
& 8.81 / 8.28
& 7.21 / 7.06 \\

Qwen-Image-2511
& 2.79 / 3.23
& 3.42 / 3.34
& 2.55 / 2.44
& 7.82 / 7.73
& 5.02 / 6.22
& 7.41 / 7.26
& 4.84 / 5.04 \\

DreamOmni2
& 2.62 / 3.09
& 4.03 / 3.74
& 2.25 / 2.39
& 5.55 / 5.59
& 4.53 / 6.13
& 7.97 / 7.75
& 4.49 / 4.78 \\

FLUX.1 Kontext
& 2.60 / 3.20
& 4.23 / 3.86
& 2.33 / 2.43
& 5.12 / 5.20
& 4.45 / 6.06
& 8.01 / 7.82
& 4.46 / 4.76 \\

Qwen-Image-2509
& 2.36 / 2.72
& 2.87 / 2.83
& 2.38 / 2.24
& 7.03 / 6.76
& 3.63 / 4.99
& 4.35 / 5.02
& 3.77 / 4.09 \\
\bottomrule
\end{tabular}
}
\end{table}

\subsection{Cross Judge Correlation}
\label{sec:human_correlation_appendix}

To validate our VLM-based evaluation protocol, we measure Pearson's linear
correlation coefficient and Spearman's rank-order correlation
coefficient between each VLM judge and human ratings on a
$168$-image human-evaluation subset.
We additionally report Human--Human agreement as a reference.
As shown in \autoref{tab:judge_human_full}, GPT-5 and Gemini 2.5 Flash both
show positive correlations with human evaluations across all criteria,
achieving PLCC/SRCC values of $0.78/0.75$ and $0.74/0.71$,
respectively, for the Average Score.
In particular, GPT-5 achieves $0.76/0.72$ for Visual Instruction
Adherence, comparable to the Human--Human agreement of $0.75/0.74$.
To examine whether this agreement is specific to the two closed-source
judges, we additionally evaluate
Qwen3-VL-32B~\citep{bai2025qwen3vl} as an open-source judge.
Qwen3-VL-32B achieves $0.73/0.70$ on the Average Score, comparable to
Gemini.
These results indicate that the VIF-Bench evaluation protocol is not
specific to a particular proprietary VLM and can also be instantiated with
an open-source judge.

\begin{table}[ht]
\centering
\small
\caption{
Correlation between human and VLM judges.
Each entry reports PLCC / SRCC on the 168-image human-evaluation subset.
Qwen3-VL-32B is included as an open-source judge.
}
% \vspace{-0.1in}
\label{tab:judge_human_full}
\scalebox{1.0}{
\begin{tabular}{lcccc}
\toprule
Metric
& Gemini--Human
& GPT-5--Human
& Qwen3-VL--Human
& Human--Human \\
\midrule
Text Instr. Follow.
& 0.69 / 0.67
& 0.72 / 0.69
& 0.64 / 0.69
& 0.63 / 0.63 \\

Reference Consist.
& 0.73 / 0.72
& 0.73 / 0.71
& 0.67 / 0.69
& 0.75 / 0.78 \\

Visual Instr. Adher.
& 0.53 / 0.60
& 0.76 / 0.72
& 0.61 / 0.63
& 0.75 / 0.74 \\

Visual Instr. Clean.
& 0.70 / 0.72
& 0.71 / 0.72
& 0.71 / 0.72
& 0.76 / 0.78 \\

Scene Coherence
& 0.48 / 0.45
& 0.53 / 0.55
& 0.51 / 0.50
& 0.58 / 0.59 \\

Visual Quality
& 0.49 / 0.57
& 0.57 / 0.63
& 0.49 / 0.59
& 0.57 / 0.60 \\

Average
& 0.74 / 0.71
& 0.78 / 0.75
& 0.73 / 0.70
& 0.80 / 0.78 \\
\bottomrule
\end{tabular}
}
\end{table}

\subsection{Consistency with Aesthetic Predictors}
\label{app:aesthetic_predictor}

To further validate the Visual Quality evaluation, we compare the
VLM-based scores with conventional image-quality predictors.
Specifically, we compute correlations with LAION Aesthetic Predictor (AP) v1 and v2~\citep{laion2022aesthentic} over 168 generated images.
\autoref{tab:aesthetic_correlation} reports the PLCC and SRCC for GPT-5, Gemini, Qwen3-VL-32B, and human Visual Quality ratings.
All three VLM judges show moderate positive correlations with both
aesthetic predictors, indicating that their Visual Quality assessments
are broadly consistent with conventional image-level quality measures.
At the same time, GPT-5 Visual Quality scores achieve a PLCC/SRCC of
0.57/0.63 with human ratings, as reported in
\autoref{tab:judge_human_full}.
These results provide complementary evidence that the VLM-based Visual
Quality criterion captures perceptual quality in a manner consistent
with both dedicated aesthetic predictors and human judgments.

\begin{table}[ht]
\centering
\small
\caption{
Correlation between Visual Quality scores and LAION Aesthetic Predictor (AP)~\citep{laion2022aesthentic} v1/v2 over 168 generated images.
}
\label{tab:aesthetic_correlation}
\scalebox{1.0}{
\begin{tabular}{lccccc}
\toprule
Score source
& $n$
& AP v1 PLCC
& AP v1 SRCC
& AP v2 PLCC
& AP v2 SRCC \\
\midrule
GPT-5 Visual Quality
& 168 & 0.566 & 0.506 & 0.680 & 0.598 \\
Gemini Visual Quality
& 168 & 0.472 & 0.382 & 0.551 & 0.468 \\
Qwen3-VL-32B Visual Quality
& 168 & 0.540 & 0.438 & 0.640 & 0.524 \\
Human Visual Quality
& 168 & 0.415 & 0.380 & 0.497 & 0.435 \\
\bottomrule
\end{tabular}
}
\end{table}

\subsection{Consistency with Detector-based Spatial Judgments}
\label{app:geometric_validation}

We further examine whether the VLM-based evaluation of visual-instruction
adherence can be complemented by specialized geometric metrics.
In particular, following the layout instructions, we use Grounding DINO~\citep{liu2023grounding}
to localize the target subjects in 168 generated images and compare the
detected regions with their instructed layout regions.
We consider two geometric measures.
\emph{Bounding-box IoU} measures the intersection-over-union between the
detected subject box and its target layout box.
\emph{Center-in-region rate} measures whether the center of the detected
subject falls inside the instructed region.
We then compute the correlation between these geometric measures and
human ratings.

As shown in \autoref{tab:geometric_correlation}, the specialized
geometric metrics exhibit moderate correlations with human judgments.
For comparison, GPT-5 achieves a PLCC / SRCC of 0.76 / 0.72 for Visual
Instruction Adherence, close to the human--human agreement of
0.75 / 0.74.
These results indicate that the VLM judge can capture spatial
instruction-following behavior at least competitively with these
detector-based alternatives, while remaining applicable to visual
instructions beyond bounding-box layouts.

\begin{table}[ht]
\centering
\small
\caption{
Correlation of detector-based geometric metrics~\citep{liu2023grounding} with human ratings
over 168 layout examples.
}
\label{tab:geometric_correlation}
\scalebox{1.0}{
\begin{tabular}{lccc}
\toprule
Geometric metric & $n$ & PLCC & SRCC \\
\midrule
Bounding-box IoU      & 168 & 0.357 & 0.435 \\
Center-in-region rate & 168 & 0.508 & 0.551 \\
\bottomrule
\end{tabular}
}
\end{table}

\subsection{Diversity under Repeated Generation}
\label{app:diversity}

The primary VIF-Bench evaluation measures whether a generated image
satisfies its references and instructions, but does not directly measure
variation across repeated generations.
To examine this aspect, we select 12 tasks and generate five outputs for
each task and model using identical references and instructions.
Following \citet{kim2025das}, we measure diversity among repeated outputs using the mean pairwise CLIP~\citep{radford2021clip}
distance and mean pairwise LPIPS~\citep{zhang2018lpips}.
The tasks are grouped by the number of major visual instructions,
denoted by $N_{\mathrm{VI}}$.

Overall, variation across repeated generations tends to increase as the
number of visual constraints grows, while the Average Score decreases.
For Nano Banana Pro, for example, CLIP diversity increases from 0.078 at
$N_{\mathrm{VI}}=1$ to 0.216 at $N_{\mathrm{VI}}=4$, while the Average
Score decreases from 7.74 to 6.48.
GPT-Image-1.5 remains comparatively stable for
$N_{\mathrm{VI}}\leq3$, but at $N_{\mathrm{VI}}=4$ its diversity
increases and its Average Score drops substantially.
Importantly, greater variation is not necessarily desirable in the
strongly constrained setting considered by VIF-Bench.
When all reference and visual-instruction constraints are satisfied,
repeated outputs are expected to remain within a relatively restricted
set of valid solutions.
As task complexity increases, different generations may instead violate
different subsets of the constraints, causing the outputs to diverge in
different directions.
The observed increase in diversity may therefore reflect generation
instability rather than useful creative variation.
Distinguishing diversity that preserves instruction fidelity from
variation caused by inconsistent constraint satisfaction remains an
important direction for future evaluation.

\begin{table}[ht]
\centering
\small
\caption{
Diversity across repeated generations from identical references and
instructions.
We generate five outputs for each task.
Higher CLIP and LPIPS distances indicate greater variation.
}
\label{tab:generation_diversity}
\scalebox{1.0}{
\begin{tabular}{lccccc}
\toprule
Model
& $N_{\mathrm{VI}}$
& \#Tasks
& CLIP div. $\uparrow$
& LPIPS div. $\uparrow$
& Avg. Score $\uparrow$ \\
\midrule
GPT-Image-1.5   & 1 & 3 & 0.066 & 0.412 & 8.55 \\
GPT-Image-1.5   & 2 & 3 & 0.042 & 0.381 & 8.58 \\
GPT-Image-1.5   & 3 & 3 & 0.079 & 0.438 & 8.49 \\
GPT-Image-1.5   & 4 & 3 & 0.124 & 0.505 & 6.03 \\
\midrule
Nano Banana Pro & 1 & 3 & 0.078 & 0.481 & 7.74 \\
Nano Banana Pro & 2 & 3 & 0.142 & 0.542 & 7.40 \\
Nano Banana Pro & 3 & 3 & 0.173 & 0.567 & 7.04 \\
Nano Banana Pro & 4 & 3 & 0.216 & 0.578 & 6.48 \\
\bottomrule
\end{tabular}
}
\end{table}

% \clearpage
\section{Extended Related Work}
\label{sec:ext_related}

\begin{table*}[ht]
\centering
\small
\caption{
Further comparison among major benchmarks for reference-based image generation,
editing, and visual-instruction following.
% \textsuperscript{$\dagger$}DreamOmni3 does not release these statistics.
}
\label{tab:benchmarks_full}
% \vspace{-0.1in}
\scalebox{0.8}{
\begin{tabular}{lccccl}
\toprule
\textbf{Benchmark}
& \textbf{\#Size}
& \textbf{\#Refs}
& \textbf{\#VIs}
& \textbf{Reference--VI Conflict}
& \textbf{Metrics} \\
\midrule

\multicolumn{6}{l}{\textit{Without visual instructions}} \\

EditBench~\citep{wang2023editbench}
& 240
& 1
& --
& \textcolor{cb_red}{\XSolidBrush}
& CLIP~\citep{radford2021clip} \\

EditVal~\citep{basu2023editval}
& 648
& 1
& --
& \textcolor{cb_red}{\XSolidBrush}
& CLIP, VLM, manual \\

EmuEdit~\citep{sheynin2024emuedit}
& 3{,}055
& 1
& --
& \textcolor{cb_red}{\XSolidBrush}
& L1, CLIP, DINO~\citep{caron2021dino} \\

MagicBrush~\citep{zhang2023magicbrush}
& 1{,}053
& 1
& --
& \textcolor{cb_red}{\XSolidBrush}
& L1, L2, CLIP, DINO \\

AnyEdit~\citep{yu2025anyedit}
& 1{,}250
& 1
& --
& \textcolor{cb_red}{\XSolidBrush}
& L1, CLIP, DINO \\

I2EBench~\citep{ma2024i2ebench}
& 2{,}240
& 1
& --
& \textcolor{cb_red}{\XSolidBrush}
& GPT~\citep{openai2023gpt4} \\

ImgEdit-Bench~\citep{ye2025imgedit}
& 811
& 1
& --
& \textcolor{cb_red}{\XSolidBrush}
& GPT (3 dim.), Fake Det.~\citep{xu2024fakeshield} \\

DreamBooth~\citep{ruiz2022dreambooth}
& 75
& 1
& --
& \textcolor{cb_red}{\XSolidBrush}
& CLIP, DINO \\

OmniContext~\citep{wu2025omnigen2}
& 400
& 3
& --
& \textcolor{cb_red}{\XSolidBrush}
& GPT (3 dim.) \\

DreamOmni2~\citep{xia2025dreamomni2}
& 319
& 4
& --
& \textcolor{cb_red}{\XSolidBrush}
& Gemini, Doubao~\citep{bytedance2025doubao} \\

MultiBanana~\citep{oshima2026multibanana}
& 3{,}769
& 8
& --
& \textcolor{cb_red}{\XSolidBrush}
& GPT, Gemini (5 dim.) \\

\midrule
\multicolumn{6}{l}{\textbf{\textit{With visual instructions}}} \\

MultiRef~\citep{chen2025multiref}
& 1{,}990
& 6
& 1
& \textcolor{cb_red}{\XSolidBrush}
& GPT (3 dim), modality-specific metrics \\

VIBE~\citep{zhang2026vibe-benchmark}
& 1{,}034
& 1
& 1
& \textcolor{cb_red}{\XSolidBrush}
& GPT (3 dim.) \\

DreamOmni3~\citep{xia2025dreamomni3}
& 731
& 4
& $2^\dagger$
& \textcolor{cb_red}{\XSolidBrush}
& Gemini, Doubao~\citep{bytedance2025doubao} \\

\bluecell{\textbf{VIF-Bench (Ours)}}
& \bluecell{1{,}241}
& \bluecell{7}
& \bluecell{6}
& \bluecell{\textbf{\textcolor{cb_green}{\Checkmark}}}
& \bluecell{GPT, Gemini, Qwen (6 dim.)} \\

\bottomrule
\end{tabular}
}
\end{table*}

\paragraph{Benchmark for image editing.}
Instruction-based image editing can be viewed as one of the earliest forms of reference-conditioned image generation, where the source image serves as a dense reference whose content must be preserved except for the instructed change. Benchmarks in this line, including MagicBrush~\citep{zhang2023magicbrush}, EMU-Edit~\citep{sheynin2024emuedit}, SmartEdit~\citep{huang2024smartedit}, I2E-Bench~\citep{ma2024i2ebench}, and ImgEdit~\citep{ye2025imgedit}, accordingly assume a single reference image and a textual edit instruction, and therefore do not address compositional multi-reference settings.
\autoref{tab:benchmarks_full} shows further comparison with prior benchmarks.

\paragraph{Visual-instruction editing.}
While early instruction-based image editing methods largely rely on natural-language prompts~\citep{hertz2023prompttoprompt, miyake2025negative-prompt-inversion}, recent work has explored richer visual instruction channels that allow users to specify edit intent more directly and unambiguously. 
VIBE~\citep{zhang2026vibe-benchmark} also emphasizes this motivation, describing visual instructions as a “more natural and efficient interaction paradigm” for communicating spatial and structural intent.
Exemplar- and demonstration-based methods condition editing on reference images or before--after visual examples, enabling users to convey appearance or stylistic changes that are difficult to describe in text~\citep{yang2022paint, nguyen2023visual}.
Spatially grounded conditioning further exposes low-level visual controls, such as edges, depth maps, poses, segmentation maps, and sketches, to constrain the edited image's geometry and layout~\citep {zhang2023controlnet}. 
In parallel, interactive manipulation methods allow users to specify geometric changes through points or drag handles, providing fine-grained control over object pose, shape, and position~\citep{pan2023draggan, shi2023dragdiffusion, mou2024dragondiffusion, ling2024freedrag}. 
Complementary to these model-centric efforts, recent datasets and benchmarks such as MagicBrush~\citep{zhang2023magicbrush} and RealEdit~\citep{sushko2025realedit} study instruction-guided editing in more realistic settings, highlighting the gap between synthetic editing tasks and practical user intent.

\paragraph{Test-time Scaling and Agents for Multimodal Generation.}
Test-time scaling (TTS), which improves model capabilities by allocating additional computation at inference time, has its roots in the development of reasoning in large language models~\citep{kojima2022large, snell2024scaling, matsutani2026rl}. 
This paradigm has recently been extended to image and video generation, where a growing body of work improves generation quality and human preferences~\citep{furuta2024improving, onoda2026multiaxis} by scaling inference-time computation without updating model parameters~\citep{yeh2024sampling, zhao2026latsearch, saini2026cachedsearch}. 
Image generation agents are a type of TTS, and they combine prompt adaptation \citep{hao2023promptist,datta2024prompt_expansion}, tool orchestration \citep{shen2023hugginggpt,wang2024genartist}, and visual feedback \citep{yang2025idea2img} to improve model outputs.
GEMS~\citep{he2026gems} integrates iterative generation with trajectory memory and reusable skills.

\paragraph{Instruction-following evaluation in LLMs.}
IFEval~\citep{zhou2023ifeval}, FollowBench~\citep{jiang-etal-2024-followbench}, Multi-IF~\citep{he2024multiif}, Multi-Instructions~\citep{harada2025multi-instructions}, and related~\citep{liu2024lost_middle, laban2026llms} all observe that LLMs ``game'' scoring by satisfying instructions in letter rather than spirit. 

% \clearpage
\section{Limitations and Future Work}
\label{sec:ext_limitations}

\paragraph{Synthetic-reference bias and source coverage.}
The reference-image pool combines real images from LAION-5B~\citep{schuhmann2022laion}, DreamOmni2~\citep{xia2025dreamomni2}, and DreamBooth~\citep{ruiz2022dreambooth} with synthetic images generated by Nano Banana and GPT-Image-1. 
Prior work evaluating this style of curated mixed-source pool has confirmed that statistical bias remains low and that the resulting datasets are reliable for benchmarking purposes~\citep{oshima2026multibanana}, so we do not consider this a blocking risk. 
As a coverage improvement, however, future versions of VIF-Bench would benefit from broadening the synthetic side to include outputs from additional generators such as Qwen-Image~\citep{wu2025qwenimagetechnicalreport} and FLUX~\citep{labs2025flux1kontext}, so that the source distribution is no longer concentrated on the GPT-Image and Nano Banana families.

\paragraph{Visual-instruction diversity.}
The five kinds of visual instructions in VIF-Bench---layout boxes, wind/light arrows, 3D orientation pyramids, and pose references---follow the design of prior work~\citep{zhang2026vibe-benchmark}, where the effectiveness of each instruction kind has already been validated. 
The space of plausible visual instructions, however, is wider than the five we currently cover. 
On the more structured end, future extensions could automatically generate additional instruction kinds, e.g., human skeletons or part-segmentation maps; on the more freeform end, they could include hand-drawn instructions such as rough scribbles, doodles, or arrow sketches that better reflect how a human user might communicate compositional intent in practice. Evaluating models against this broader range would test not only whether they follow the well-defined visual-instruction kinds we use here, but whether they generalize to the full distribution of compositional cues a creative user is likely to draw.

\paragraph{Extension to video generation.}
The evaluation philosophy of VIF-Bench naturally extends to subject-driven video generation~\citep{googledeepmind2025veo3_1, openai2024sora, chen2025videoalchemist, zhang2026multirefcompass}, where multiple references and heterogeneous visual instructions must be satisfied consistently over time without leaving residual instruction marks. 
In videos, constraints such as layout, orientation, pose, lighting, wind, and arrow-specified motion must remain coherent across frames, motivating frame-level evaluation of visual instruction adherence and both spatial and temporal evaluation of scene coherence. 
Moreover, past observations or visual memories maintained by video-generation world models~\citep{xiao2025worldmem, oshima2026worldpack} could be treated as multiple references, enabling evaluation of long-term subject and state consistency as well as memory--instruction interactions when past visual context conflicts with current instructions.

\clearpage
\section{Prompts}
\label{sec:prompts}

This appendix lists the full prompts used in VIF-Bench's construction and evaluation pipeline.

\subsection{Visual Instructions Proposal Prompts}
\label{sec:visual_instructions_prompt}

Below we list the four prompts used to propose visual instructions during task construction. 
A VLM (e.g., GPT-5~\citep{openai2025gpt5}) receives the sampled main reference images and the canvas dimensions $W \times H$, and returns a JSON record that is then rendered into the corresponding visual-instruction image (layout boxes, direction arrows, or 3D orientation pyramids, as shown in \autoref{fig:image_source_stats}; \textbf{Right}). 

\begin{center}
\begin{minipage}{\textwidth}
\begin{tcolorbox}[title=Layout Proposal Prompt]
You are a visual layout designer. You are given $N$ reference subject images in order:

\vspace{0.5mm}
\noindent
\{listing\}

\vspace{1.0mm}
Propose a composition on a canvas of $W \times H$ pixels that will render as ONE seamless natural image---foreground and background integrated, as if it were a single photograph or illustration. AVOID collage-like or grid-like arrangements that produce visible panel boundaries.

\vspace{1.0mm}
For EACH reference image, decide where its main subject should be placed on the canvas.

\vspace{1.0mm}
Return ONLY a JSON object with this exact shape (no markdown, no commentary):

\vspace{0.5mm}
\noindent
\{ \\
\quad ``canvas'': [$W$, $H$], \\
\quad ``boxes'': [ \\
\quad\quad \{``index'': 0, ``box'': [$x_\text{min}$, $y_\text{min}$, $x_\text{max}$, $y_\text{max}$]\}, \\
\quad\quad \{``index'': 1, ``box'': [$x_\text{min}$, $y_\text{min}$, $x_\text{max}$, $y_\text{max}$]\}, \\
\quad\quad \ldots \\
\quad ] \\
\}

\vspace{1.0mm}
\noindent
Rules:

\vspace{0.5mm}
\noindent
1. Box format is [$x_\text{min}$, $y_\text{min}$, $x_\text{max}$, $y_\text{max}$] normalized to $[0.0, 1.0]$ (top-left origin). \\
2. \texttt{index} MUST reference the image ordering above (0-based), so each box is tied to a specific source image. \\
3. Include one box per input image, in index order. \\
4. NO BOX MAY SPAN THE ENTIRE IMAGE, regardless of how many boxes there are. Every box must leave visible margin from the canvas edges: $x_\text{min} \geq 0.05$, $y_\text{min} \geq 0.05$, $x_\text{max} \leq 0.95$, $y_\text{max} \leq 0.95$. A box that covers $90\%+$ of both dimensions is forbidden even in single-subject tasks---leave background space around the subject. \\
5. Composition goal: the final image should read as a single cohesive scene. Prefer overlapping placements, foreground-background depth cues, and natural spatial relationships (e.g., smaller objects in front of larger subjects, subjects sharing a common ground plane). \\
6. AVOID tight grids, equal-sized side-by-side panels, or any arrangement that would make the final image look like separate pictures pasted together. Leave room for background / negative space to connect the subjects. \\
7. Boxes may overlap when the composition calls for it.
\end{tcolorbox}
\end{minipage}
\end{center}

\begin{center}
\begin{minipage}{\textwidth}
\begin{tcolorbox}[title=Arrow Proposal Prompt]
You are a visual scene designer. You are given $N$ reference subject images that will be composed into one scene on a $W \times H$ canvas:

\vspace{0.5mm}
\noindent
\{listing\}

\vspace{1.0mm}
Your job is to propose ONE directional arrow PER requested kind, indicating the direction of an environmental factor across the canvas. Each arrow has a TAIL (start) and a HEAD (end), in normalized canvas coordinates (top-left origin, range $[0.0, 1.0]$).

\vspace{1.0mm}
\noindent
Requested arrows:

\vspace{0.5mm}
\noindent
- \texttt{wind}: Wind arrow. The TAIL is where the wind originates, the HEAD is the direction the wind blows TOWARDS. Choose a direction that is physically plausible for the scene (e.g., across the subjects, not aimed straight into the ground).

\vspace{1.0mm}
Return ONLY a JSON object with this exact shape (no markdown, no commentary):

\vspace{0.5mm}
\noindent
\{ \\
\quad ``canvas'': [$W$, $H$], \\
\quad ``arrows'': [ \\
\quad\quad \{``label'': ``wind'', ``start'': [$x$, $y$], ``end'': [$x$, $y$]\} \\
\quad ] \\
\}

\vspace{1.0mm}
\noindent
Rules:

\vspace{0.5mm}
\noindent
1. Coordinates are normalized to $[0.0, 1.0]$; $(0, 0)$ = top-left, $(1, 1)$ = bottom-right. \\
2. Each arrow MUST have meaningful length (Euclidean distance between start and end $\geq 0.4$ in normalized units), so the direction is unambiguous when rendered. \\
3. Keep BOTH endpoints inside the canvas with at least $0.05$ margin from any edge (i.e., $0.05 \leq x, y \leq 0.95$). \\
4. The arrow must clearly indicate direction, not be a near-zero-length blob. \\
5. Include exactly one arrow per requested kind, in the order listed above. \\
6. Do NOT add arrows for kinds that were not requested.
\end{tcolorbox}
\end{minipage}
\end{center}

\begin{center}
\begin{minipage}{\textwidth}
\begin{tcolorbox}[title=Light Arrow Proposal Prompt]
You are a visual scene designer. You are given $N$ reference subject images that will be composed into one scene on a $W \times H$ canvas:

\vspace{0.5mm}
\noindent
\{listing\}

\vspace{1.0mm}
Your job is to propose ONE directional arrow PER requested kind, indicating the direction of an environmental factor across the canvas. Each arrow has a TAIL (start) and a HEAD (end), in normalized canvas coordinates (top-left origin, range $[0.0, 1.0]$).

\vspace{1.0mm}
\noindent
Requested arrows:

\vspace{0.5mm}
\noindent
- \texttt{light}: Light arrow. The TAIL is at the light source, the HEAD points TOWARDS where the light falls on the subjects. Choose a key-light direction consistent with the scene (typically from above, off-axis, never straight up from the ground).
\vspace{1.0mm}
Return ONLY a JSON object with this exact shape (no markdown, no commentary):

\vspace{0.5mm}
\noindent
\{ \\
\quad ``canvas'': [$W$, $H$], \\
\quad ``arrows'': [ \\
\quad\quad \{``label'': ``light'', ``start'': [$x$, $y$], ``end'': [$x$, $y$]\} \\
\quad ] \\
\}

\vspace{1.0mm}
\noindent
Rules:

\vspace{0.5mm}
\noindent
1. Coordinates are normalized to $[0.0, 1.0]$; $(0, 0)$ = top-left, $(1, 1)$ = bottom-right. \\
2. Each arrow MUST have meaningful length (Euclidean distance between start and end $\geq 0.4$ in normalized units), so the direction is unambiguous when rendered. \\
3. Keep BOTH endpoints inside the canvas with at least $0.05$ margin from any edge (i.e., $0.05 \leq x, y \leq 0.95$). \\
4. The arrow must clearly indicate direction, not be a near-zero-length blob. \\
5. Include exactly one arrow per requested kind, in the order listed above. \\
6. Do NOT add arrows for kinds that were not requested.
\end{tcolorbox}
\end{minipage}
\end{center}

\begin{center}
\begin{minipage}{\textwidth}
\begin{tcolorbox}[title=Orientation Proposal Prompt]
You are a visual scene designer. The composition uses a $W \times H$ canvas. For each subject listed below, decide which 3D direction that subject should FACE in the final image.

\vspace{1.0mm}
\noindent
Subjects requiring a facing direction:

\vspace{0.5mm}
\noindent
\{subject\_descriptor\_lines\}

\vspace{0.5mm}
\noindent
\{layout\_preview\_sentence\}

\vspace{1.0mm}
Coordinate convention (right-handed, camera at origin looking down $-Z$; $+Z$ = OUT of the screen toward the viewer):

\vspace{0.5mm}
\noindent
- yaw $= 0^\circ$ $\rightarrow$ subject faces TOWARD the viewer ($+Z$) \\
- yaw $= +90^\circ$ $\rightarrow$ subject faces SCREEN-RIGHT ($+X$) \\
- yaw $= 180^\circ$ $\rightarrow$ subject faces AWAY from viewer ($-Z$) \\
- yaw $= -90^\circ$ $\rightarrow$ subject faces SCREEN-LEFT ($-X$) \\
- diagonals $\rightarrow$ $3/4$ views (e.g., yaw $= 45^\circ$ = front-right $3/4$ from the camera) \\
- pitch $= 0^\circ$ $\rightarrow$ level \\
- pitch $= +30^\circ$ $\rightarrow$ looking UP (head tilted up) \\
- pitch $= -30^\circ$ $\rightarrow$ looking DOWN

\vspace{1.0mm}
Pick yaw + pitch that are physically and aesthetically plausible for the subject given the composition:

\vspace{0.5mm}
\noindent
- People commonly face slightly toward the camera or toward another subject. \\
- Animals often face toward food, another subject, or where action is. \\
- Outdoor objects (cars, buildings) face along their natural axis (front of the vehicle, entrance of the building).

\vspace{1.0mm}
\noindent
CONSTRAINT (the rendered pyramid must be visually distinguishable):

\vspace{0.5mm}
\noindent
- $|\text{yaw}| < 60^\circ$ AND $|\text{pitch}| < 60^\circ$ \\
- $|\text{yaw}| + |\text{pitch}| \leq 75^\circ$ \\
- NOT ($|\text{yaw}| < 15^\circ$ AND $|\text{pitch}| < 15^\circ$): at least one of $|\text{yaw}|, |\text{pitch}|$ must be $\geq 15^\circ$. The (yaw, pitch) region too close to $(0, 0)$ renders as a flat square and is excluded.

\vspace{0.5mm}
\noindent
Pick angles inside this region. Examples of valid (yaw, pitch): $(+15, 0)$, $(0, +15)$, $(+30, -15)$, $(-30, +15)$, $(+45, -30)$, $(-45, +30)$, $(0, -45)$, $(+15, -45)$, $(-15, +45)$, $(+45, +15)$.

\vspace{1.0mm}
Return ONLY a JSON object with this exact shape (no markdown, no commentary):

\vspace{0.5mm}
\noindent
\{ \\
\quad ``canvas'': [$W$, $H$], \\
\quad ``cones'': [ \\
\quad\quad \{ \\
\quad\quad\quad ``main\_index'': $\langle$int$\rangle$, \\
\quad\quad\quad ``yaw\_deg'': $\langle$number; $|\text{yaw}| < 60$$\rangle$, \\
\quad\quad\quad ``pitch\_deg'': $\langle$number; $|\text{pitch}| < 60$$\rangle$, \\
\quad\quad\quad ``rationale'': ``$\langle$one short sentence explaining the choice$\rangle$'' \\
\quad\quad \} \\
\quad ] \\
\}

\vspace{1.0mm}
\noindent
Rules:

\vspace{0.5mm}
\noindent
1. Include exactly ONE cone entry per subject listed above. \\
2. Respect the CONSTRAINT above. Do not output (yaw, pitch) close to $(0, 0)$ or with magnitude $\geq 60^\circ$ in any single axis. \\
3. Do NOT propose apex, length, or base radius---those are fixed by the renderer.
\end{tcolorbox}
\end{minipage}
\end{center}

\clearpage
\subsection{Judge Prompt}
\label{sec:judge_prompt}

The judge prompt is fed to each judge VLM (Gemini 2.5 Flash and GPT-5) in the canonical input order: (1) the $N$ reference images, (2) the visual instructions, (3) the textual instruction given to the generator, and (4) the generated output image. The judge produces a free-form ``Reasoning'' segment followed by one numerical score per criterion on a 1--10 scale.

\begin{center}
\begin{minipage}{\textwidth}

\begin{tcolorbox}[title=VIF-Bench Judge Prompt]
You are a STRICT evaluator for a multi-reference image generation system.
You will be given reference images, visual instructions, the textual instruction given to the generator, and the generated output image.

\vspace{0.5mm}
Reference Images: \{reference\_image\_files\} \\
Visual Instructions: \{visual\_instruction\_files\} \\
Instruction: \{instruction\} \\
Generated Image: \{generated\_image\_file\}

\vspace{1.0mm}
Your task is to evaluate the generated image from six independent perspectives, each on a 10-point scale.
For each criterion, follow the hard caps strictly: a violation triggers a fixed score ceiling regardless of the rest of the image.

\vspace{1.0mm}
\noindent
1. Text Instruction Following

\vspace{0.5mm}
Evaluate how faithfully the generated image follows the textual requirements in the instruction that are not visual instructions. 
This includes global requirements that shape the whole scene (the overall style, the background setting) and per-subject requirements that bind a specific attribute to a specific main subject. 
For text references, only the text itself must appear in the output, without the background of the text reference image.
If any of these requirements is ignored, the score must not exceed 5.
If a per-subject attribute is applied to the wrong subject, or the required overall style is not applied at all, the score must not exceed 4.
If the background of a text reference is carried into the output, the score must not exceed 5.

\vspace{1.0mm}
\noindent
2. Reference Consistency

\vspace{0.5mm}
Evaluate subject fidelity along identity, shape, texture, and font dimensions.
If a subject is missing from the output or replaced with a different subject, the score must be 1.
If any single subject fails recognizable detail matching with its reference, the score must not exceed 6.

\vspace{1.0mm}
\noindent
3. Vision Instruction Adherence (STRICT)

\vspace{0.5mm}
Evaluate adherence to ALL visual instructions provided in this task, which may include: (a) layout boxes — each subject must occupy its assigned colored box; (b) 3D orientation pyramids — each subject must face the direction indicated by its pyramid (front = square base, back = apex); (c) wind/light arrows — global wind/illumination in the scene must follow the cyan/yellow arrow direction;
(d) pose references — the designated person must take the pose shown in the pose reference.
Apply the following hard caps. When multiple visual instructions are present, the WORST violation across them determines the cap.
If subjects are swapped across the colored bounding boxes (wrong color-to-subject mapping), OR a pose reference is applied to the wrong person, the score must be 1.
If any subject is clearly misplaced relative to its assigned box, OR a subject's facing direction is clearly opposite (e.g., left/right flipped, or front/back reversed) to its orientation pyramid, OR the rendered wind/light direction is clearly opposite to its arrow, OR the designated person's pose clearly does not match the pose reference, the score must not exceed 3.
If any subject visibly protrudes outside its assigned box, OR the facing direction / wind / light / pose only loosely matches the visual instruction (recognizable but inaccurate), the score must not exceed 6.

\vspace{1.0mm}
\noindent
4. Visual Instruction Cleanliness (STRICT)

\vspace{0.5mm}
Evaluate whether any visual-instruction mark has been carried over into the
generated image, or newly drawn on top of it. Visual-instruction marks include:
rectangular layout frames, 3D orientation pyramids, direction arrows
(wind/light), and any other recognizable instructional overlay.
Count one "complete mark" as one recognizable instance of any of the above
(e.g., one full box, one full pyramid, one full arrow). Partial or faint
remnants that are still recognizable count as marks; only fully integrated,
unrecognizable traces are exempt.
If any 

\end{tcolorbox}

\end{minipage}
\end{center}

\clearpage
\begin{center}
\begin{minipage}{\textwidth}

\begin{tcolorbox}

recognizable mark of any kind remains in the output, the score must not exceed 6.
If exactly one complete mark is preserved or drawn into the output, the score must be 4.
If two or more complete marks remain (of any kind, in any combination), the score must be 1.

\vspace{1.0mm}
\noindent
5. Scene Coherence

\vspace{0.5mm}
Evaluate foreground-background integration and semantic coherence between subjects.
If the result reads as an obvious collage of pasted parts, the score must not exceed 3.
If there is a style or lighting mismatch between subjects and background, the score must not exceed 5.
If subjects from semantically distinct situational contexts are blended in a pixel-smooth but incoherent way, the score must not exceed 4.

\vspace{1.0mm}
\noindent
6. Visual Quality

\vspace{0.5mm}
Evaluate the overall perceptual quality of the image, independent of instruction following.
Assess whether the image is visually appealing and aesthetically coherent.
This criterion has no caps tied to other criteria.

\vspace{1.0mm}
Each of the six scores must be evaluated independently. Do not force any score to be tied to or capped by another score except as stated above.

\vspace{1.0mm}
First, explain the reasoning, then present the final assessment. \\
Start the reasoning with Reasoning: .

\vspace{0.5mm}
After explaining the reasoning, present the final assessment in the format:

\vspace{0.5mm}
\noindent
Text Instruction Following: $\langle$A number from 1 to 10$\rangle$. \\
Reference Consistency: $\langle$A number from 1 to 10$\rangle$. \\
Vision Instruction Adherence: $\langle$A number from 1 to 10$\rangle$. \\
Visual Instruction Cleanliness: $\langle$A number from 1 to 10$\rangle$. \\
Scene Coherence: $\langle$A number from 1 to 10$\rangle$. \\
Visual Quality: $\langle$A number from 1 to 10$\rangle$.

\end{tcolorbox}
\end{minipage}
\end{center}

\end{document}